\documentclass[12pt,onecolumn]{IEEEtran}
\usepackage{setspace}
\usepackage[section]{placeins}
\usepackage{graphicx,subfigure}
\usepackage{amsmath}
\usepackage{amsfonts}
\usepackage{caption2}
\usepackage{amssymb}
\usepackage{varwidth}
\usepackage[rightbars,color]{changebar}
\usepackage{epsfig}
\usepackage{multirow}
\usepackage{booktabs}
\usepackage[linesnumbered,ruled,vlined]{algorithm2e}
\usepackage{array}
\usepackage{rotating}
\usepackage{epsfig,amsfonts,multirow}
\usepackage{changebar}
\newcommand{\punt}[1]{}

\begin{document}
\title{Learning Multi-Scale Deep Features for High-Resolution Satellite Image Classification}
\author{~Qingshan~Liu,~\IEEEmembership{Senior~Member,~IEEE}, Renlong~Hang, ~Huihui~Song, ~Zhi~Li
\thanks{Q. Liu, H. Song and Z. Li are with the Jiangsu Key Laboratory of Big Data Analysis Technology,
the School of Information and Control, Nanjing University of Information Science and Technology,
Nanjing 210044, China (qsliu$@$nuist.edu.cn).

R. Hang is with the Jiangsu Key Laboratory of Big Data Analysis Technology, the School of Atmospheric Science and School of Information and Control, Nanjing University of Information Science and Technology, Nanjing 210044, China (renlong\_hang$@$163.com).}}

\maketitle

\begin{spacing}{2}
\begin{abstract}
In this paper, we propose a multi-scale deep feature learning method for high-resolution satellite image classification. Specifically, we firstly warp the original satellite image into multiple different scales. The images in each scale are employed to train a deep convolutional neural network (DCNN). However, simultaneously training multiple DCNNs is time-consuming. To address this issue, we explore DCNN with spatial pyramid pooling (SPP-net). Since different SPP-nets have the same number of parameters, which share the identical initial values, and only fine-tuning the parameters in fully-connected layers ensures the effectiveness of each network, thereby greatly accelerating the training process. Then, the multi-scale satellite images are fed into their corresponding SPP-nets respectively to extract multi-scale deep features. Finally, a multiple kernel learning method is developed to automatically learn the optimal combination of such features. Experiments on two difficult datasets show that the proposed method achieves favorable performance compared to other state-of-the-art methods.
\end{abstract}
\IEEEpeerreviewmaketitle

\section{Introduction}\label{Introduction}
Remote sensing image classification has been an active research topic in the past few decades, and most of the existing works primarily focus on pixel-wise classification, which assigns label information to each pixel in a multi-spectral or hyper-spectral image \cite{benediktsson2005}\cite{melgani2004}\cite{plaza2009}\cite{fauvel2013}. Although significant progress has been made in this area, pixels are not enough for the entire image understanding because they have few semantic meanings \cite{cheng2015}. With the development of imaging techniques, large amounts of high spatial resolution satellite images become available \cite{dai2011}\cite{xia2010}\cite{yang2011}, which opens new possibilities in remote sensing image analysis and classification.

\begin{figure}[htp]
  \centering
  \includegraphics[scale = 0.7]{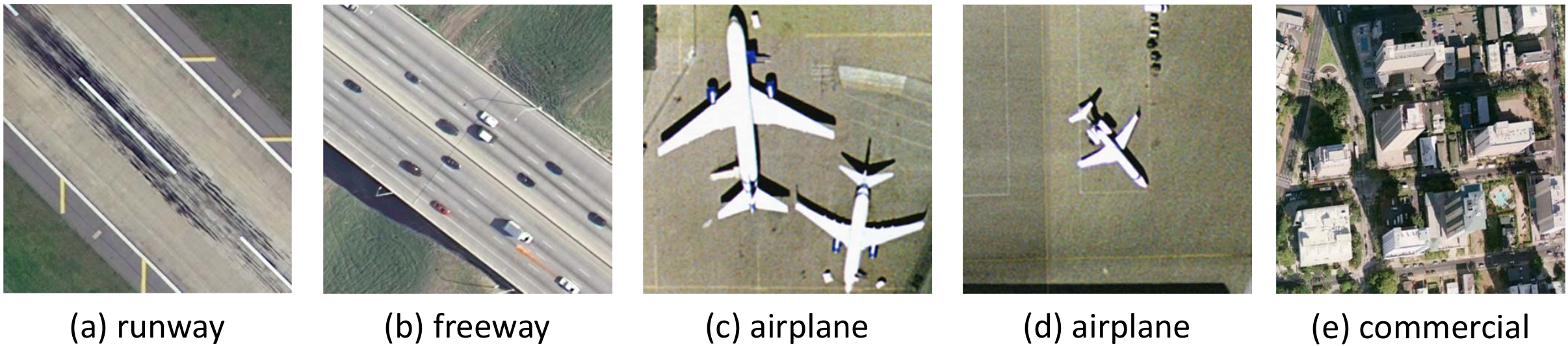}\\
  \caption{A few examples of high-resolution satellite images.}\label{satellite_image}
\end{figure}

However, satellite images with high spatial resolution pose many challenging issues in image classification. First, the enhanced resolution brings more details, thus simple low-level features (e.g., intensity and textures) widely used in the case of low-resolution images are insufficient in capturing efficiently discriminative information \cite{xia2010}. For instance, Figs.$~$\ref{satellite_image} (a) and (b) have similar color and texture features, but they belong to different categories (i.e., runway and freeway), which can be discriminated by adding the car information. Second, objects in the same type of scene might have different scales and orientations \cite{sheng2012}. As shown by
Figs.$~$\ref{satellite_image} (c) and (d), the airplane in (d) is much smaller than that in (c), and their orientations are also different. Besides, high-resolution satellite images often consist of many different semantic classes, which makes further classification more difficult \cite{cheriyadat2014}. Taking Fig.$~$\ref{satellite_image} (e) for example, the commercial scene comprises roads, buildings, trees, parking lots, etc. Thus, developing efficient feature representations is critical for solving these issues.

There have two popular feature representation models that are successfully used in satellite image classification. One is the Bag of Visual Words (BOVW) model \cite{sivic2003}\cite{csurka2004}\cite{fei2005}, which generally includes three steps: 1) extracting man made visual features, such as SIFT \cite{lowe2004} and HOG \cite{dalal2005} descriptors; 2) clustering features to form visual words (clustering centers) by using k-means or other clustering methods; 3) mapping visual features to the closest word and generating a mid-level feature representation by word histograms.
This model and its variants have been investigated in satellite image classification \cite{yang2008}\cite{cheriyadat2014}. However, it is an orderless collection of local descriptors, regardless of spatial information. To overcome this drawback, spatial pyramid matching (SPM) method was proposed in \cite{lazebnik2006}, in which the image is firstly partitioned into increasingly fine sub-regions and then histograms of local features are extracted inside each sub-region. Since satellite imagery generally does not have an absolute reference frame, the relative spatial arrangement of the image elements becomes very important. Accordingly, the authors in \cite{yang2011} proposed  the spatial pyramid co-occurrence, which characterizes both the photometric and geometric information of an image.
Unlike dividing the image into uniform cells in \cite{yang2011} and \cite{lazebnik2006}, the authors in
\cite{jiang2012} proposed the randomized spatial partition to characterize various image layout.

Feature representation based on sparse coding (SC) is the other popular method for scene classification \cite{wright2010}\cite{boureau2010}. Its basic idea is that the original signal can be sparsely reconstructed with respect to some fixed bases (dictionary) and the selected bases (training samples) are enforced into as few categories as possible. In \cite{dai2011}, a two-layer SC was proposed for satellite image classification. In \cite{sheng2012}, Sheng \textit{et al.} proposed to use SC to generate three mid-level representations based on SIFT, local ternary pattern histogram fourier (LTP-HF) and colour histogram features, respectively. Recently, an unsupervised dictionary learning method has been proposed in \cite{cheriyadat2014} which achieves favorable performance in satellite image classification.

Although these methods have achieved promising results in satellite image classification, there still exist some shortcomings. For the BOVW models, a key step is how to extract low-level visual features. This process is generally hand-crafted and heavily depends on experience and domain knowledge of designers. For the SC models, they can be considered as a single-layer feature learning architecture, which automatically selects a few vectors from a large pool of possible bases to encode an input signal \cite{yu2011}\cite{bengio2013}.
As discussed in \cite{deng2014}, the shallow architectures have shown effectiveness in solving many simple or well-constrained problems, but their limited modeling and representational power are insufficient in complex scene cases like the high resolution satellite images. Besides, SC focuses on searching for sparse representation of the original images, which may lose helpful discriminative information for the subsequent supervised classification.

Recently, deep learning, especially DCNN, has attracted increasing attention in natural image processing \cite{bengio2013}. The core idea is to hierarchically learn high-level semantic features without human interactions. In 2012, Krizhevsky \textit{et al.} designed a DCNN architecture based on two GPUs with multiple convolutional and fully-connected layers \cite{krizhevsky2012}. This architecture achieved excellent classification results on the ImageNet 2012 Large Scale Visual Recognition Challenge. Afterwards, a large amount of works about DCNN sprang up \cite{sermanet2014}\cite{howard2013}\cite{zeiler2014}\cite{chatfield2014}\cite{simonyan2014}\cite{he2014}.
In \cite{he2014}, He \textit{et al.} proposed SPP-net to solve the size constraint problem of input images, which exists in most DCNN architectures.
Benefiting from spatial pyramid pooling, SPP-net can be trained faster and achieves higher performance than DCNN.
In the field of remote sensing image processing, to the best of our knowledge, there are only a few literatures about DCNN \cite{yue2015} mainly because it is difficult to acquire a large amount of training samples.

In this paper, we employ SPP-net to automatically extract multi-scale deep features of high-resolution satellite images.
As shown in Figs.$~$\ref{satellite_image} (c) and (d), the scales of objects in satellite images often vary. Traditional DCNNs are not able to sufficiently explore this information, because they can only extract the deep features of images from a pre-defined scale (e.g., $224\times224$). We therefore attempt to construct multiple DCNNs with different input scales to address this issue. However, it is well known that training a deep model costs much time, not to mention training multiple models simultaneously. Benefiting from spatial pyramid pooling, SPP-net can generate a fixed-length representation regardless of image size/scale. In other words, SPP-nets with different input image scales can exactly share the same weight parameters \cite{he2015}. Besides, for each SPP-net, fine-tuning the parameters in fully-connected layers ensures an efficient network, thus greatly accelerating the training process. Therefore, we choose SPP-net as our basic deep model.
Because of the large numbers of parameters and scarce of training samples, the SPP-net inevitably poses overfitting problem. We therefore take advantage of the training results using ImageNet dataset. Afterwards, we use the trained SPP-nets to extract multi-scale deep features. In the subsequent classification process, to optimally fuse such features, we develop a multiple kernel learning method.

The major contributions of this paper are summarized as follows:
\begin{enumerate}
\item To effectively apply SPP-nets, a pre-training method using ImageNet dataset is proposed to address the small sample size issue in high-resolution satellite image classification.
  \item Since the scales of objects in satellite images often vary, multiple SPP-nets are successfully applied to capture such information. Each SPP-net corresponds to one-scale images.
  \item Multi-scale deep features represent multi-level abstract information of satellite images.
        In order to automatically combine these features for the subsequent classification, we propose a multiple kernel learning method. Numerous experimental results certify the effectiveness and superiority of the proposed method.
\end{enumerate}

The rest of this paper is organized as follows. In Section II, we present the proposed method in detail, including the SPP-net architecture, the training method of the architecture and the multiple kernel learning framework. The experiments are reported in Section III, followed by the conclusion in Section IV.
\begin{figure}
  \centering
  \includegraphics[scale = 0.6]{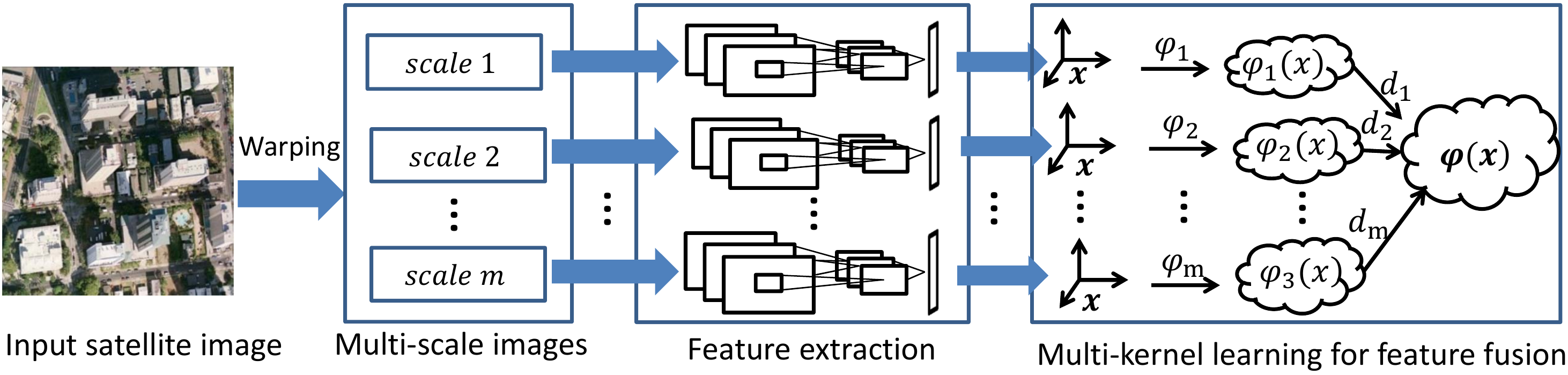}\\
  \caption{Flowchart of the proposed method.}\label{overview}
\end{figure}
\section{Methodology}

The flowchart of the proposed method is shown in Fig.$~$\ref{overview}. The whole procedure includes three steps: 1) warping the original satellite images into multiple scales;
2) multi-scale deep features extraction using multiple SPP-nets; 3) multi-scale deep features fusion via a multi-kernel learning method.
In the following subsections, we introduce the last two steps in detail.

\subsection{SPP-net architecture}
\begin{figure}
  \centering
  \includegraphics[scale = 0.48]{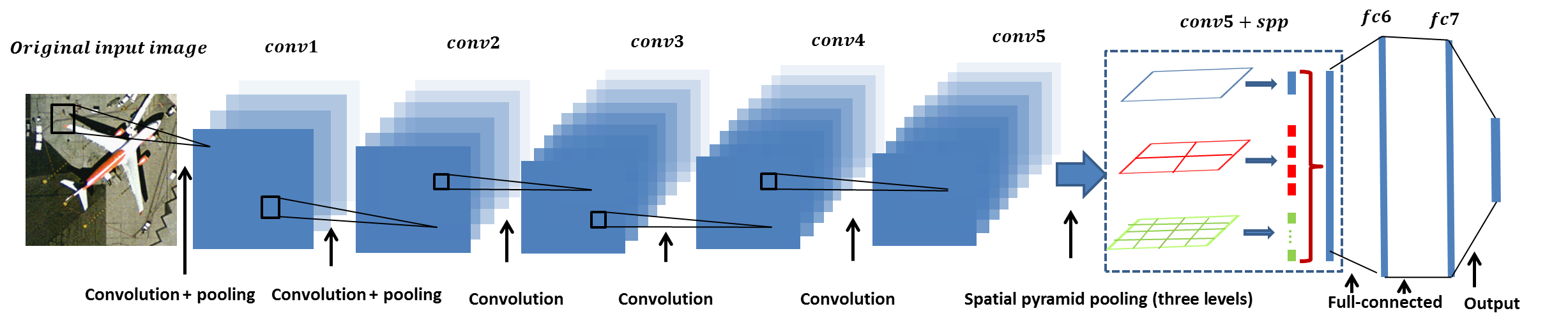}\\
  \caption{The architecture of SPP-net.}\label{spp_net}
\end{figure}
SPP-net was firstly proposed in \cite{he2014} to address the size issue of input images. Here, we use it to automatically learn multi-scale deep features of high-resolution satellite images. Specifically, we combine the prevalent seven-layer architecture in \cite{krizhevsky2012} with spatial pyramid pooling (SPP). The designed architecture is shown in Fig.$~$\ref{spp_net}. The network contains five successive convolutional layers and two fully-connected layers. The first two convolutional layers are followed by max-pooling operators. They operate in a sliding-window manner and output feature maps representing the spatial layout of the responses. Before the first fully-connected layer, SPP is exploited to pool the features from the last convolutional layer. Similar to SPM \cite{lazebnik2006}, we partition the feature maps into increasingly fine sub-regions, and pool the responses inside each sub-region (throughout this paper, we use max pooling). Assume the size of each feature map after the last convolutional layer is $a\times a$ pixels and each feature map is partitioned into $n\times n$ sub-regions. Then, SPP can be considered as convolution operators in a sliding-widow manner with window size $\textit{win} = \lceil a/n\rceil $ and stride $\textit{str} = \lfloor a/n\rfloor $ , where $\lceil \cdot\rceil$ and $\lfloor \cdot\rfloor $ denote ceiling and floor operators, respectively. Fig.\ref{spp_net} demonstrates a three-level SPP configuration by setting $n\times n$ as $1\times 1$, $2\times2$ and $4\times4$, respectively. The final output of SPP is to concatenate these three-level pooling results into a vector. This simple pooling operator largely reduces the number of parameters needed to be trained between the last convolution layer and the first fully-connected layer. Thus, it is faster to train SPP-net than the traditional DCNNs. Besides, SPP extracts multi-resolution information from the last convolutional layer, which improves the final classification results. Despite the varying sizes of input images, which leads to the varying sizes of feature maps at each convolutional layer, the lengths of input vectors to the first fully-connected layer remain the same. This property ensures that the number of parameters remains unchanged. Therefore, the multiple SPP-nets are capable of sharing the same initial parameters.

\subsection{Training method}\label{training}
\begin{figure}
  \centering
  \includegraphics[scale = 0.6]{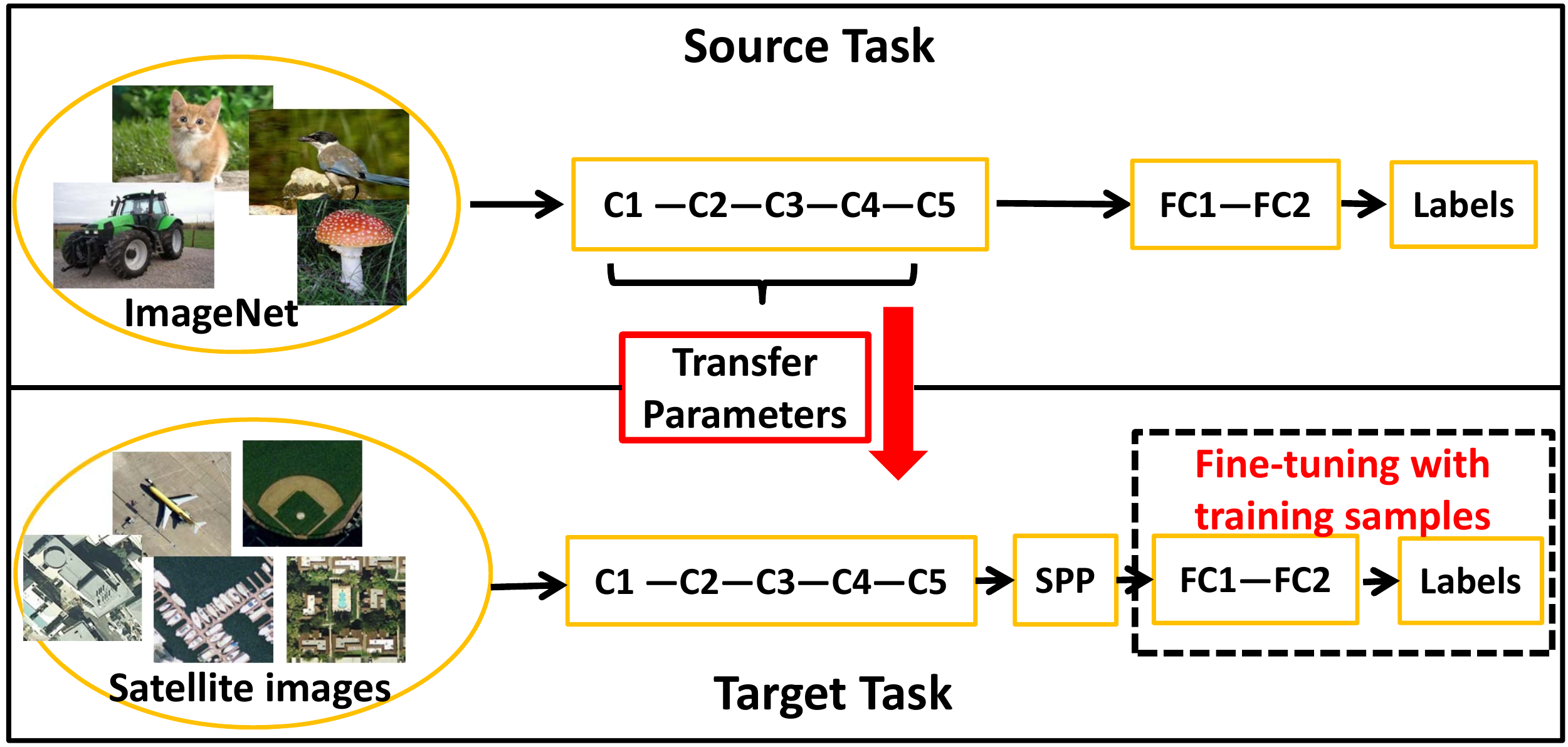}\\
  \caption{The detailed training process of our method.}\label{training_method}
\end{figure}
The above network contains more than 30 millions of parameters. Training such a network needs large amounts of samples. A canonical dataset widely used in the DCNN architectures is ImageNet, consisting of millions of images. However, only hundreds of samples are available for high-resolution satellite image classification, which is far less than ImageNet. This is the main reason why DCNN has not been successfully used in the remote sensing image processing. The most intuitional method to enlarge the number of training samples is via cropping and flipping operators \cite{krizhevsky2012}\cite{he2014}. However, it is still far from enough to train an efficient network. Recently, the authors in \cite{oquab2014} proposed to transfer image representations learned with DCNN on large datasets to other visual recognition tasks with limited training data. Motivated by this work, we propose to firstly pre-train the network in \cite{krizhevsky2012} using auxiliary ImageNet 2012 dataset (Source task), and then fine-tune our SPP-nets by employing the training samples from satellite images (Target task).

The training procedure of source task is carried out via the open source Caffe DCNN library \cite{jia2014}. Specifically, multinomial logistic regression function is optimized using stochastic gradient descent algorithm based on the back propagation method \cite{lecun1989}. The batch size and momentum are set to 256 and 0.9, respectively. The training is regularized by weight decay of 0.0005 and dropout operators for the two fully-connected layers (dropout ratio is set to 0.5).
The initial learning rate is set to be 0.01. This value is fixed and used to update iteratively the weights. At each iteration, we calculate the classification accuracy of the validation set. When the accuracy stops increasing, we divide the learning rate by 10, and this new value is used to update the weights. The whole process is repeated until convergence.
In our experiments, the learning rate reduces 3 times prior to termination (after 370K iterations) and the weights in each layer are initialized from a zero-mean Gaussian distribution with standard deviation $\sigma = 0.01$.
After the pre-training of source task, the weight parameters learned in the five convolutional layers are then transferred to the target task and kept fixed. For the target task, we only need to fine-tune the last three layers (i.e., two fully-connected layers and output layer). The whole process is demonstrated in Fig.$~$\ref{training_method}. It is worth noting that the source task is pre-trained only once, and the source task along with the target task share the same initial parameters, which means that the learned parameters from the source task are directly transferred to the multiple SPP-nets. For each SPP-net, the parameters of fully-connected layers are fine-tuned by the training samples of satellite images while other parameters remain the same. After training the networks, the multi-scale images are fed into their corresponding networks to extract multi-scale deep features.

\subsection{Multiple kernel learning}
With the extracted multi-scale deep features, the intuitive way of integrating these features is to concatenate them into a vector. This method is based on the assumption that all features have the same contribution to the subsequent classification, which obviously is not true in most cases. Besides, the formed high-dimensional feature space not only increases the computational burden, but also induces the overfitting problem. Multiple kernel learning (MKL) has been proved to be an efficient method to combine different features for remote sensing image classification \cite{camps2006}\cite{camps2008}\cite{gu2012}\cite{gu2015}. In this paper, the extracted multi-scale deep representations can be considered as different features of an image. So we employ MKL to integrate these multi-scale features.


Assume the extracted multi-scale features are denoted as $X = \{x_{1},x_{2},\cdots,x_{N}\}$, where $x_{i}\in R^{d}$ and $N$ is the number of samples. Then we map the input space to higher space as follows:
\begin{equation}
  \phi: R^{d} \rightarrow F \qquad x \rightarrow \phi(x),
\end{equation}
where $\phi$ is a non-linear mapping function and $F$ is the corresponding mapped feature space. Via this non-linear mapping, the non-separable problem in $R$ is transformed separable in $F$. Nonetheless, directly computing $\phi$ is nontrivial, but we can calculate the dot product in the high-dimensional space via kernel trick, which can be expressed as $K(x_{i},x_{j}) = \langle\phi(x_{i}),\phi(x_{j})\rangle$, where the operator $\langle\cdot\rangle$ means inner product, and $K(\cdot)$ denotes kernel function.

Before introducing the proposed multiple kernel learning method, we first consider the binary classification problem. Given the labeled training samples $\{(x_{i},y_{i}), i = 1,2,\cdots,N\}$, where $y_{i} \in \{-1,+1\}$, support vector machine (SVM) aims at searching for a linear decision function $f(x) = \langle\omega,\phi(x)\rangle + b$ maximizing the margin. It is well known that minimizing the norm of the parameters $1/2\|\omega\|^{2}$ under the constraint $y_{i}(\langle\omega,\phi(x)\rangle + b) \geq 1$ maximizes the margin. Such a minimization of the weights provides a naturally regularized solution, which favors smooth models of optimal complexity and avoids overfitting the data. The dual problem of SVM can be written as:

\begin{equation}\label{original}
\begin{aligned}
  &\max \, W(\alpha_{i},\alpha_{j})=\sum_{i=1}^{N}\alpha_{i} - \frac{1}{2}\sum_{i=1}^{N}\sum_{j=1}^{N}\alpha_{i}\alpha_{j}y_{i}y_{j}K(x_{i},x_{j}),\\
  & s.t.\quad 0\leq\alpha_{i}\leq C, \,i=1,2,\cdots,N,\\
  &\qquad\sum_{i=1}^{N}\alpha_{i}y_{i} = 0,
\end{aligned}
\end{equation}
where $\alpha_{i}$ and $\alpha_{j}$ are Lagrange multipliers, and if $\alpha_{i}$ is nonzero, the corresponding $x_{i}$ is called support vector, which determines the decision hyperplane.

The core idea of multiple kernel learning is to replace the single kernel $K$ in Eq.~(\ref{original}) with a linear combination of $M$ kernels:
\begin{equation}\label{MKL}
  \begin{aligned}
  &K(x_{i},x_{j}) = \sum_{m=1}^{M}d_{m}K_{m}(x_{i},x_{j}),\\
  &s.t.\quad d_{m}\geq0, \quad\sum_{m=1}^{M}d_{m}=1,
  \end{aligned}
\end{equation}
where $M$ is the number of candidate basis kernels, and $d_{m}$ is the weight for the $m$-th basis kernel. In this paper, $M$ is particularly set as the number of scales, and each basis kernel exploits the features in one scale. Thus, the objective function can be rewritten as:
 \begin{equation}\label{obj}
\begin{aligned}
  &\max \, W(\alpha_{i},\alpha_{j})=\sum_{i=1}^{N}\alpha_{i} - \frac{1}{2}\sum_{i=1}^{N}\sum_{j=1}^{N}\alpha_{i}\alpha_{j}y_{i}y_{j}\sum_{m=1}^{M}d_{m}K_m(x_{i},x_{j}),\\
  &s.t.\quad\sum_{i=1}^{N}\alpha_{i}y_{i} = 0,\\
  & \qquad 0\leq\alpha_{i}\leq C, \,i=1,2,\cdots,N,\\
  & \qquad d_{m}\geq0, \quad\sum_{m=1}^{M}d_{m}=1.
\end{aligned}
\end{equation}

To simultaneously optimize the combining weights $d_{m}$, $\alpha_{i}$ and $\alpha_{j}$, we adopt the SimpleMKL algorithm, which was firstly proposed in \cite{simplemkl}. Because the objective function in (\ref{obj}) is convex and differentiable, $d_{m}$ is optimized by using gradient ascend method. The gradient equals to the derivatives of $W$:
\begin{equation}
  \frac{\partial W}{\partial d_{m}} = -\frac{1}{2}\sum_{i=1}^{N}\sum_{j=1}^{N}\alpha_{i}\alpha_{j}y_{i}y_{j}K_{m}(x_{i},x_{j}),\quad m=1,2,\cdots,M.
\end{equation}
Then, $d_{m}$ is updated as follows:
\begin{equation}
  d_{m} = d_{m}+\gamma\frac{\partial W}{\partial d_{m}},
\end{equation}
where $\gamma$ is the step length. Note that the gradient is updated only when the objective value decreases during the iterative process. This updating procedure is repeated until the stopping criterion is satisfied.

\section{Experiments}
To evaluate the proposed method, we compare it with several state-of-the-art approaches on two widely used data sets: \textit{21-Class-Land-Use} \cite{yang2011} dataset and \textit{19-Class Satellite Scene} \cite{dai2011}\cite{xia2010} dataset.

\subsection{21-Class-Land-Use dataset}
\subsubsection{\textbf{Data description}}
\begin{figure}
  \centering
  \includegraphics[scale = 0.5]{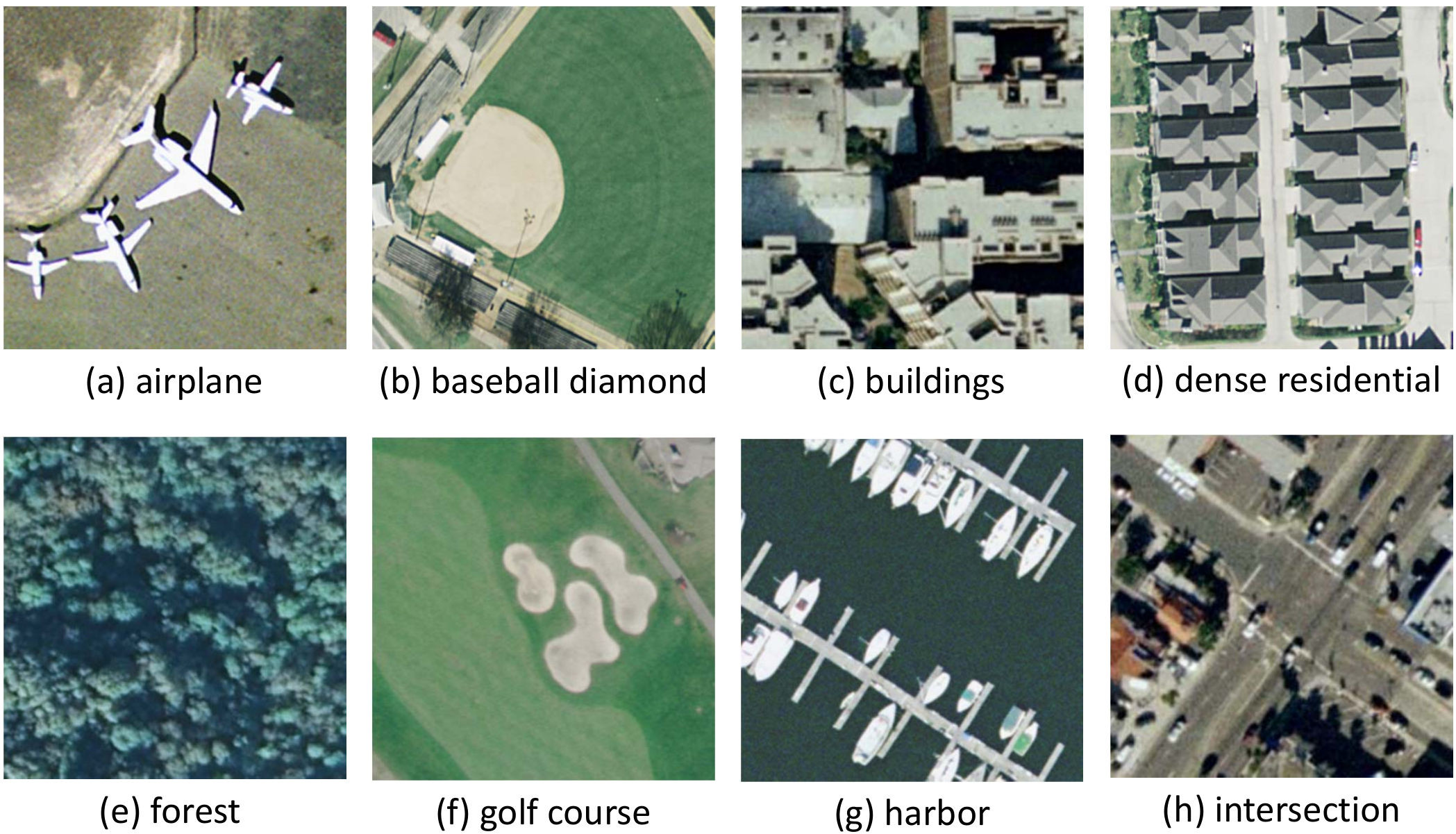}\\
  \caption{Some image examples in 21-Class-Land-Use dataset. }\label{image_examples_21}
\end{figure}

This dataset was manually extracted from aerial orthoimagery downloaded from the United States Geological Survey (USGS) National Map. It consists of 21 different land use and land cover classes, including \textit{agricultural, airplane, baseball diamond, beach, buildings, chaparral, dense residential, forest,
freeway, golf course, harbor, intersection, medium density residential, mobile home park, overpass, parking lot, river, runway, sparse residential, storage tanks} and \textit{tennis courts}. Each class contains 100 RGB images with pixel resolution of one foot (i.e., 0.3 m) and image size of $256\times256$ pixels. Fig.~\ref{image_examples_21} shows some image examples from the 21 classes.

\subsubsection{\textbf{Experimental setup}}
In each experiment, besides the original scale, the images are warped into two different scales, including $128\times128$ and $192\times192$ pixels. In multiple kernel learning step, one linear kernel is learned in each scale to map the corresponding features. For training and testing, the images in each class are randomly split into two sets. In training stage, we use the training set to fine-tune the SPP-nets and train the linear SVMs, where the SVMs are implemented using the LIBSVM package, and one-against-all strategy is adopted to address the multi-class issue. The performance of classifiers are then evaluated on the testing set. In order to reduce the effect of random selection, we repeat each algorithm on ten different training/testing split of the data set and report means and standard deviations of the obtained accuracies.

\subsubsection{\textbf{Each layer performance}}\label{each_layer_performance}

\begin{figure}
\begin{center}
   \subfigure[]{\label{21_128} \includegraphics[width=.3\linewidth]{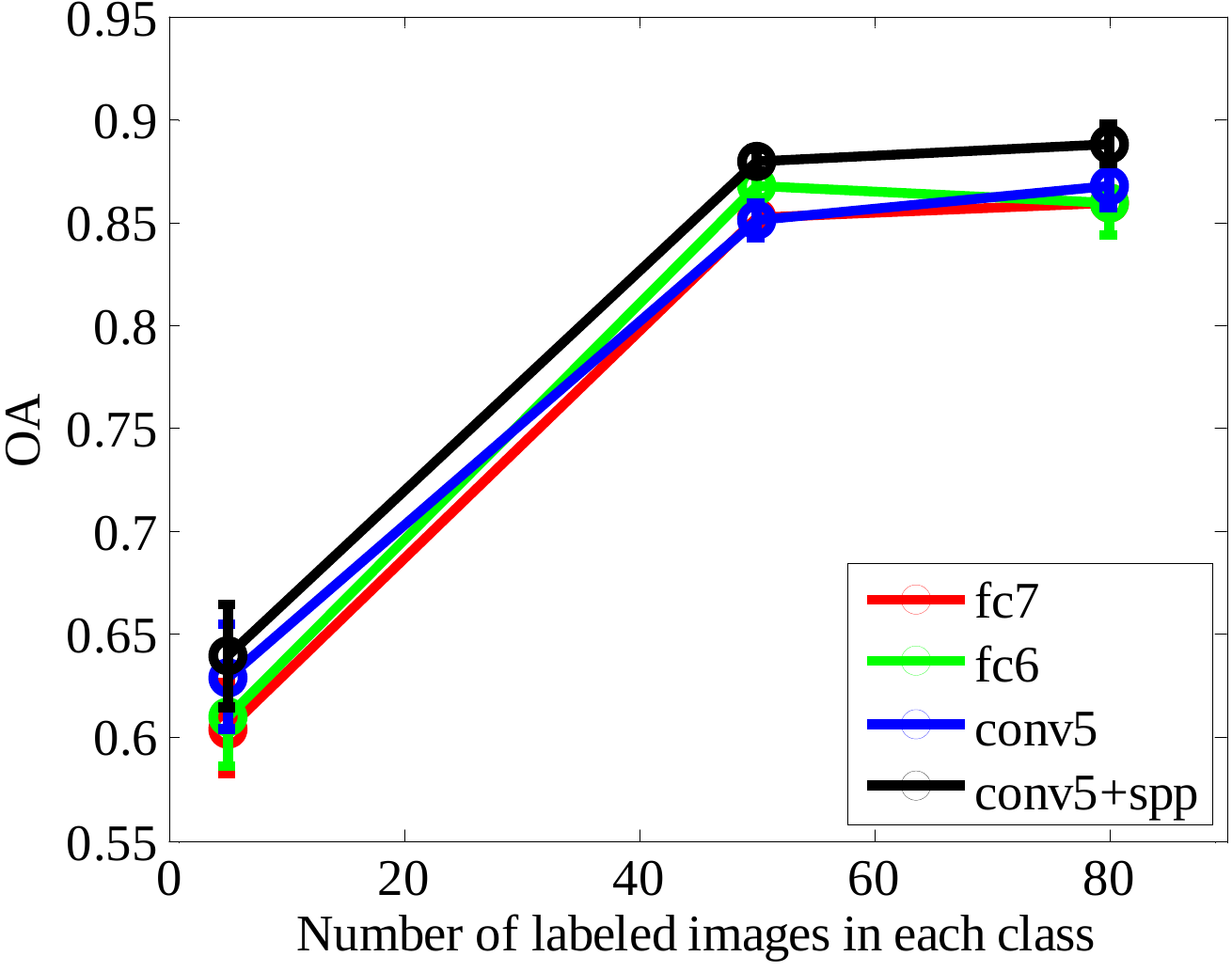}}
   \subfigure[]{\label{21_192} \includegraphics[width=.3\linewidth]{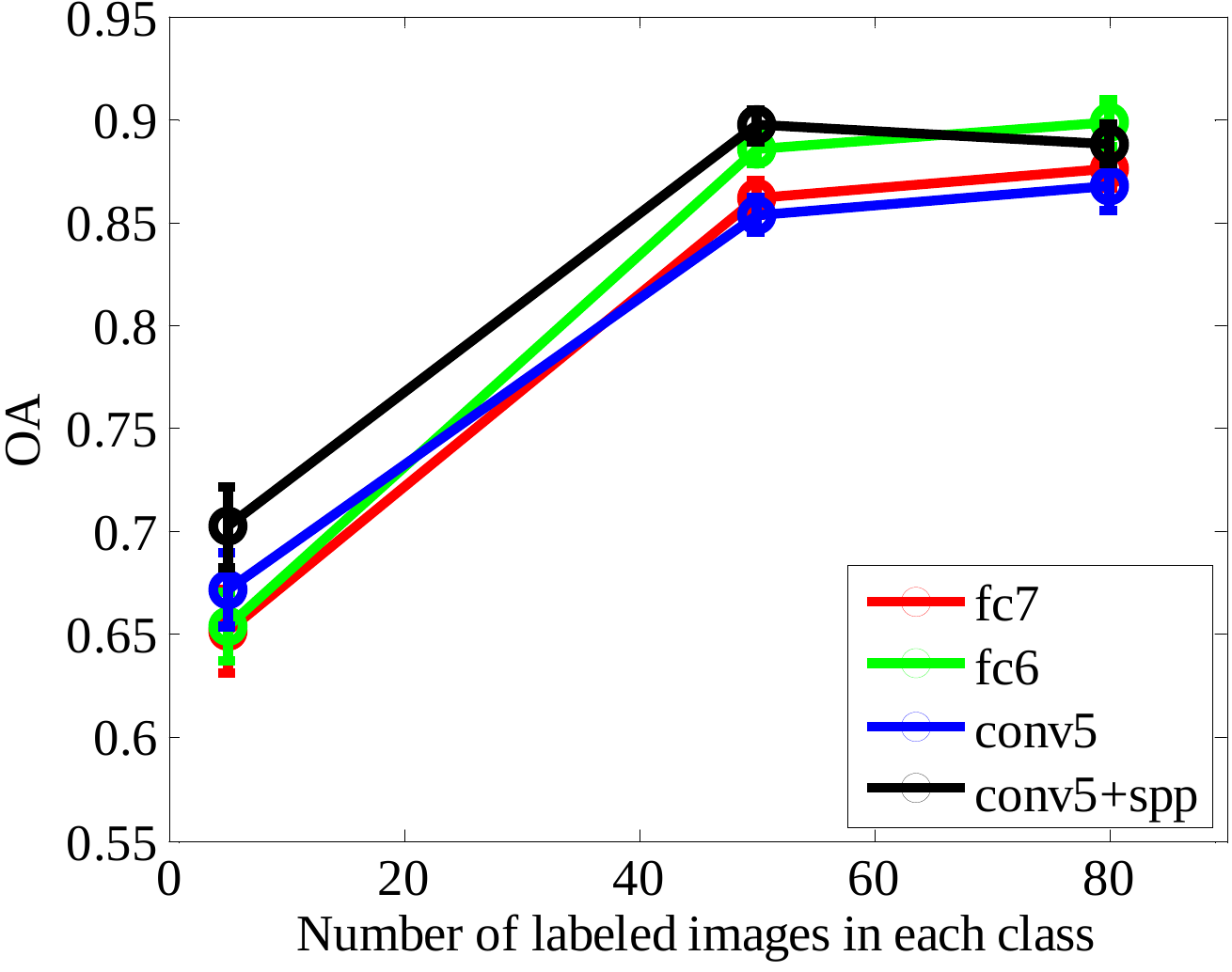}}
   \subfigure[]{\label{21_256} \includegraphics[width=.3\linewidth]{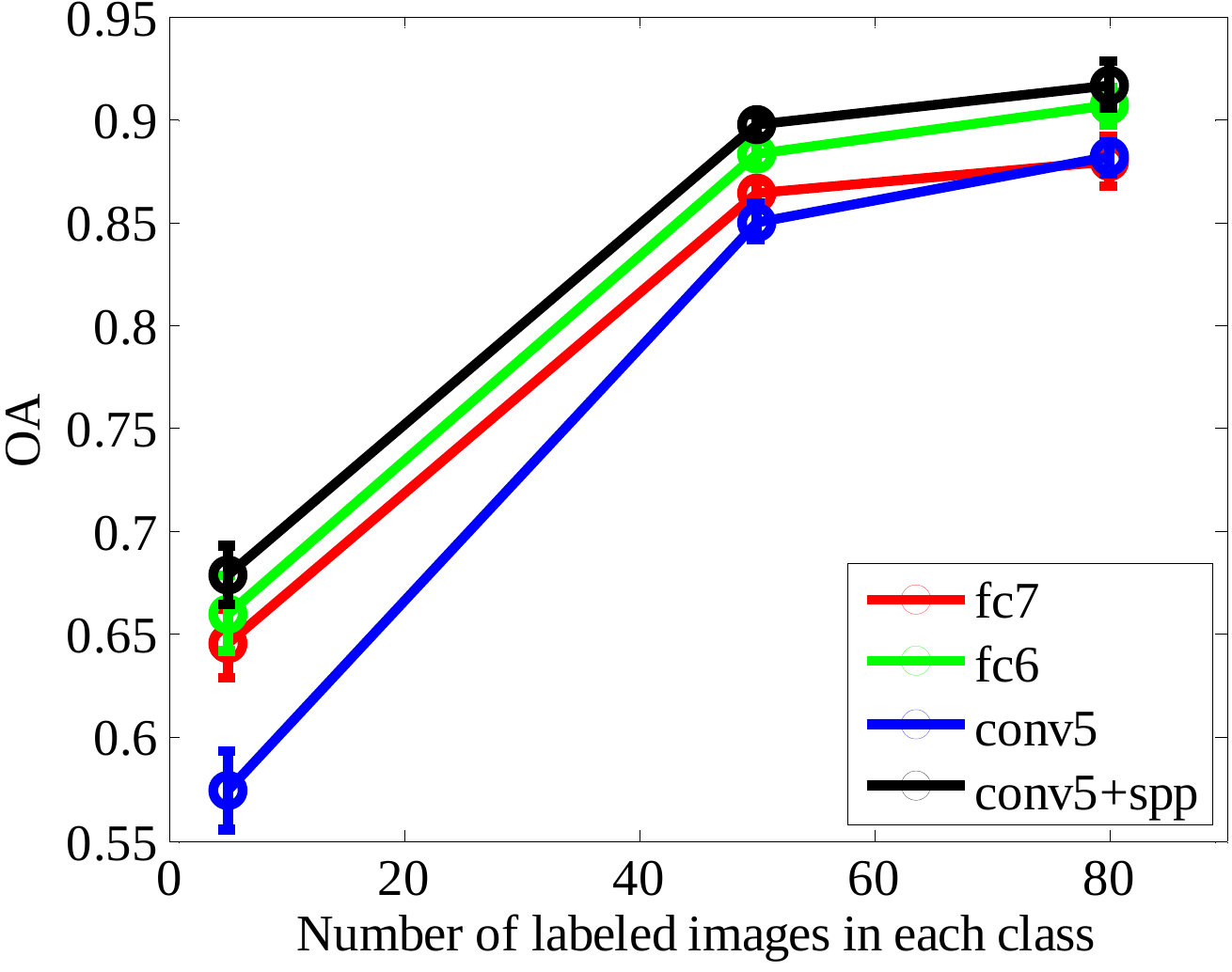}}
\end{center}
\caption{OAs and standard deviations of SPP-nets at three different scales under different number of training samples on 21-Class-Land-Use dataset. \subref{21_128} $128\times128$ scale. \subref{21_192} $192\times192$ scale. \subref{21_256} $256\times256$ scale.}
\label{each_scale_results_21}
\end{figure}

To assess which layer is the best for our task, similar to \cite{girshick2014}, we analyze and compare the results of the last four feature layers. For simplicity, we name them conv5, conv5+spp, fc6 and fc7. Fig.$~$\ref{each_scale_results_21} shows the mean overall accuracies (OAs) and standard deviations using features from different layers at three scales versus different number of training samples. From this figure, we can conclude that the OAs are improved as the number of training samples increases. Besides, fc6 is better than conv5 and fc7 in most cases, which is consistent with the conclusion in \cite{girshick2014}. However, with the spatial pyramid pooling, conv5+spp improves the performance significantly as compared to conv5, and achieves better results than fc6. Table$~$\ref{classification_results_21} demonstrates the detailed quantitative results using 5, 50 and 80 training samples from each class, respectively. The bold fonts indicate the best results with respect to different number of training samples under one scale. In common with Fig.$~$\ref{each_scale_results_21}, the features from conv5+spp layers achieve the highest accuracies in most of the cases. Thus, we use the features from conv5+spp for the subsequent multi-scale feature fusion via MKL.

\begin{table}
  \centering
   \caption{ OAs (\%) and standard deviations of SPP-nets with different layer features under different number of training samples on 21-Class-Land-Use dataset.}\label{classification_results_21}
  \begin{tabular}{|c|c|c|c|c|c|}
    \hline
    Scales & Numbers  & conv5 & conv5+spp & fc6 & fc7 \\
    \hline
    \multirow{3}{*}{$128\times128$} & 5 & $62.93\pm2.55$ & $\mathbf{63.96\pm2.52}$ & $60.99\pm2.37$ & $60.41\pm2.20$\\
    \cline{2-6}
     & 50 & $85.12\pm0.81$ & $\mathbf{87.98\pm0.50}$  & $86.79\pm0.64$ & $85.18\pm0.50$  \\
    \cline{2-6}
     & 80  & $86.81\pm1.18$ & $\mathbf{88.81\pm0.94}$  & $85.98\pm1.65$ & $85.98\pm1.65$ \\
    \hline
        \multirow{3}{*}{$192\times192$} & 5  & $67.20\pm1.81$ & $\mathbf{70.27\pm1.96}$  & $65.45\pm1.68$ & $65.14\pm2.02$\\
    \cline{2-6}
     & 50 & $85.37\pm0.82$ & $\mathbf{89.77\pm0.79}$  & $88.66\pm0.70$ & $86.25\pm0.81$  \\
    \cline{2-6}
     & 80  & $86.81\pm1.18$ & $88.81\pm0.94$   & $\mathbf{89.88\pm1.16}$ & $87.64\pm0.92$\\
    \hline
        \multirow{3}{*}{$256\times256$} & 5  & $57.40\pm1.92$ & $\mathbf{67.89\pm1.44}$  & $65.99\pm1.84$ & $64.58\pm1.64$\\
    \cline{2-6}
     & 50 & $84.99\pm0.88$ & $\mathbf{89.70\pm0.52}$  & $88.35\pm0.65$ & $86.44\pm0.52$  \\
    \cline{2-6}
     & 80  & $88.17\pm0.78$ & $\mathbf{91.67\pm1.11}$   & $90.62\pm0.89$ & $87.95\pm1.15$\\
    \hline
  \end{tabular}
\end{table}

\begin{table}
  \centering
  \caption{The detailed classification results comparison between single scale features and two different multi-scale feature fusion methods on 21-Class-Land-Use dataset. }\label{MKL_results_21_table}
  \begin{tabular}{|c|c|c|c|}
     \hline
     Numbers & 5 & 50 & 80 \\
     \hline
     Conv5+spp-128 & $63.96\pm2.52$ & $87.98\pm0.50$ & $88.81\pm0.94$ \\
     \hline
     Conv5+spp-192 & $70.27\pm1.96$ & $89.77\pm0.79$ & $88.81\pm0.94$ \\
     \hline
     Conv5+spp-256 & $67.89\pm1.44$ & $89.70\pm0.52$ & $91.67\pm1.11$ \\
     \hline
     Conv5+spp+SV & $70.57\pm2.06$ & $90.73\pm0.76$ & $91.38\pm0.46$ \\
     \hline
     Conv5+spp+MKL & $\mathbf{75.33\pm1.86}$ & $\mathbf{95.72\pm0.50}$ & $\mathbf{96.38\pm0.92}$ \\
     \hline
   \end{tabular}

\end{table}

\subsubsection{\textbf{The efficiency of MKL}}
\begin{figure}
  \centering
  \includegraphics[scale = 0.5]{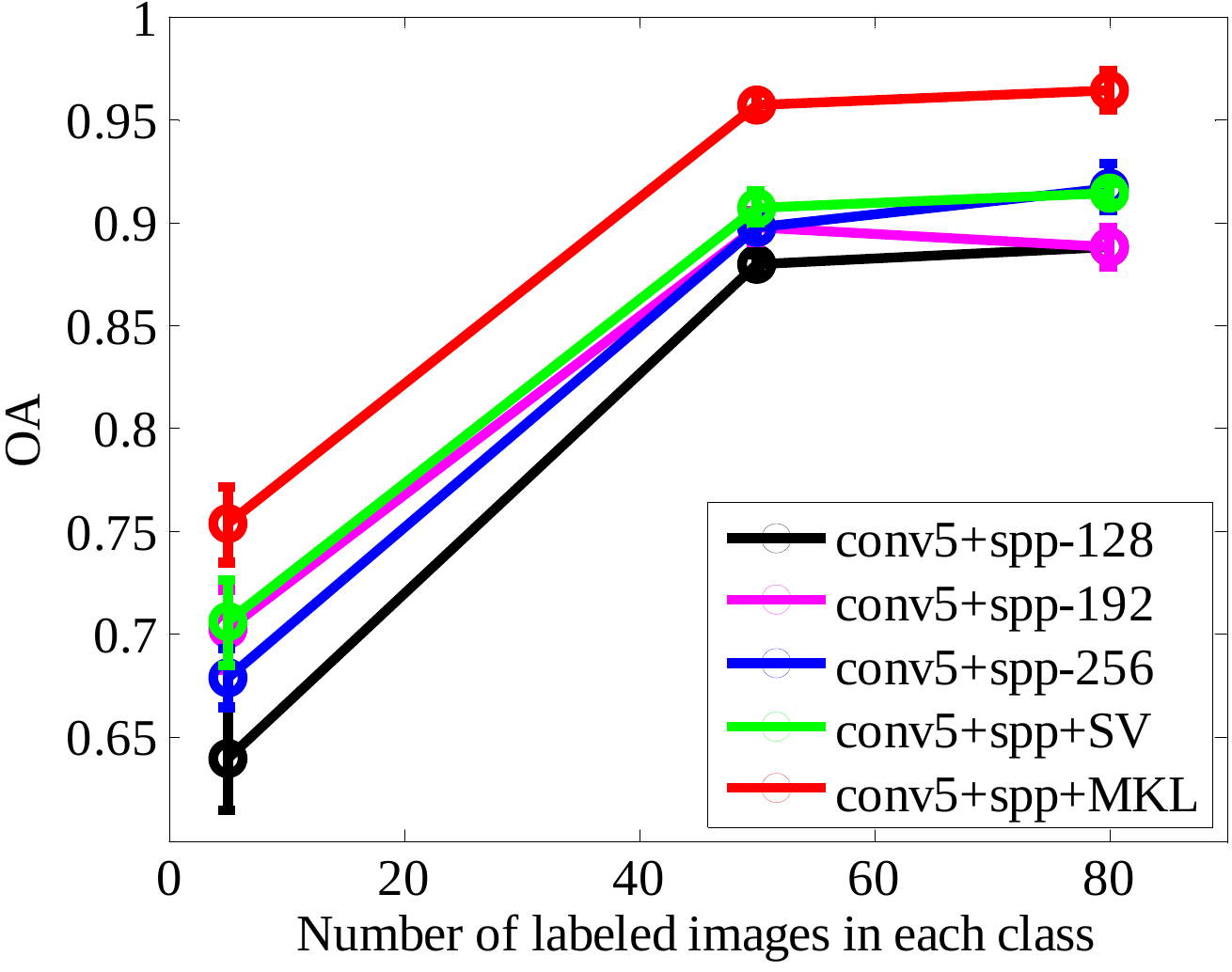}\\
  \caption{OAs and standard deviations of MKL versus SV and the single scales using conv5+spp features under different number of training samples on 21-Class-Land-Use dataset. }\label{MKL_results_21}
\end{figure}

To examine the performance of our proposed MKL fusion method, we compare it with the single scale method and the traditional fusion method, i.e., stacking the multi-scale deep features into one vector (SV). Fig.$~$\ref{MKL_results_21} demonstrates the classification results using different number of training samples with features from conv5+spp layers. From this figure, we can observe that SV method achieves higher classification accuracies than the single scale features, which can be explained that multi-scale deep features represent different abstracts of the original images and simultaneously using these complementary information thereby improves the classification results. Another obvious observation is that MKL method significantly boosts the classification results as compared to SV. The reason can be attributed to the fact that MKL automatically learns the optimal combination among multi-scale deep features, while SV simply assumes that the features in all scales play the same role. The quantitative results in Table$~$\ref{MKL_results_21_table} support the conclusions in Fig.$~$\ref{MKL_results_21}, which further confirms the efficiency of our proposed fusion method.

\subsubsection{\textbf{Comparison with the state-of-the-arts}}
\begin{table}[htp]
  \centering
  \caption{Overall classification accuracy (\%) comparison on the 21-Class-Land-Use dataset.}\label{classification_comparision_21}
  \begin{tabular}{|c|c|c|c|}
     \hline
      Numbers & 5 & 50 & 80 \\
      \hline
      SSEP \cite{yang2015} & $65.34\pm2.01$ & $-$ & $-$ \\
      \hline
      Partlets-based method \cite{cheng2015} & $-$ & $88.76\pm0.79$ & $91.33\pm1.11$ \\
      \hline
      SC+Pooling \cite{cheriyadat2014} & $-$ &$-$ & $81.67\pm1.23$ \\
      \hline
      BOVW \cite{yang2011} & $-$ &$-$ & 71.68\\
      \hline
      SPCK++ \cite{yang2011} & $-$ &$-$ & 77.38\\
      \hline
      SPMK \cite{lazebnik2006} & $-$ &$-$ & 74.00\\
      \hline
      MKL \cite{cusano2014} & $64.78\pm1.62$ & $88.68\pm1.10$ & $91.26\pm1.17$ \\
      \hline
      UFL \cite{hu2014} & $-$ &$-$ & $90.26\pm1.51$ \\
      \hline
      SPP-net & $70.27\pm1.96$ & $89.77\pm0.79$ & $91.67\pm1.11$\\
      \hline
      SPP-net+SV & $70.57\pm2.06$ & $90.73\pm0.76$ & $91.38\pm0.46$\\
      \hline
      SPP-net+MKL & $\mathbf{75.33\pm1.86}$ & $\mathbf{95.72\pm0.50}$ & $\mathbf{96.38\pm0.92}$\\
      \hline
   \end{tabular}
\end{table}

\begin{figure}[htp]
  \centering
  \includegraphics[scale = 0.5]{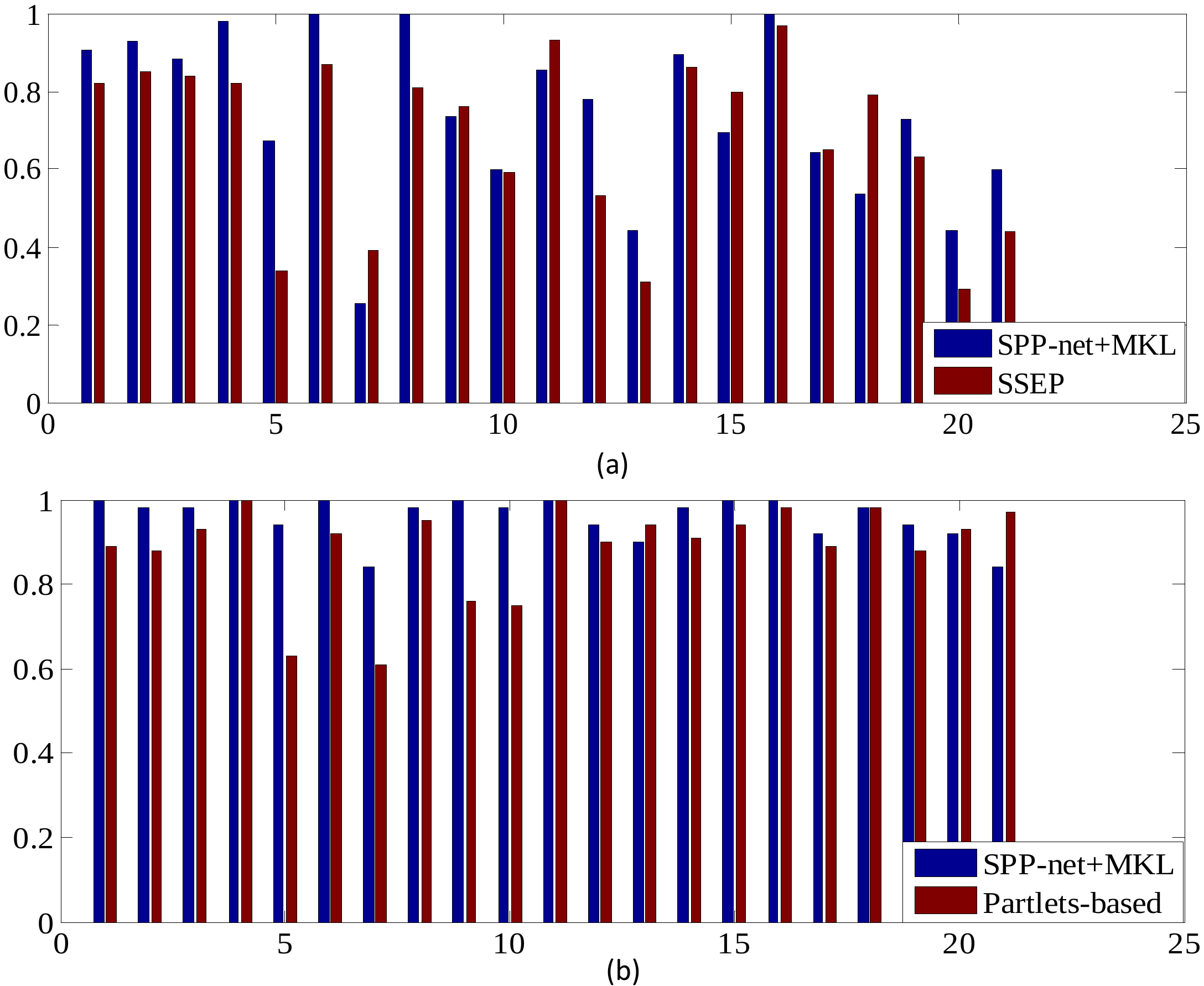}\\
  \caption{Each class accuracy comparison between two methods on 21-Class-Land-Use dataset: (a) SSEP in \cite{yang2015} and SPP-net+MKL using 5 training samples; (b) Partlets-based method in \cite{cheng2015} and SPP-net+MKL using 50 training samples.}\label{each_class_acc_21}
\end{figure}

\begin{figure}
\begin{center}
   \subfigure[]{\label{CM_21_5tr} \includegraphics[width=.45\linewidth]{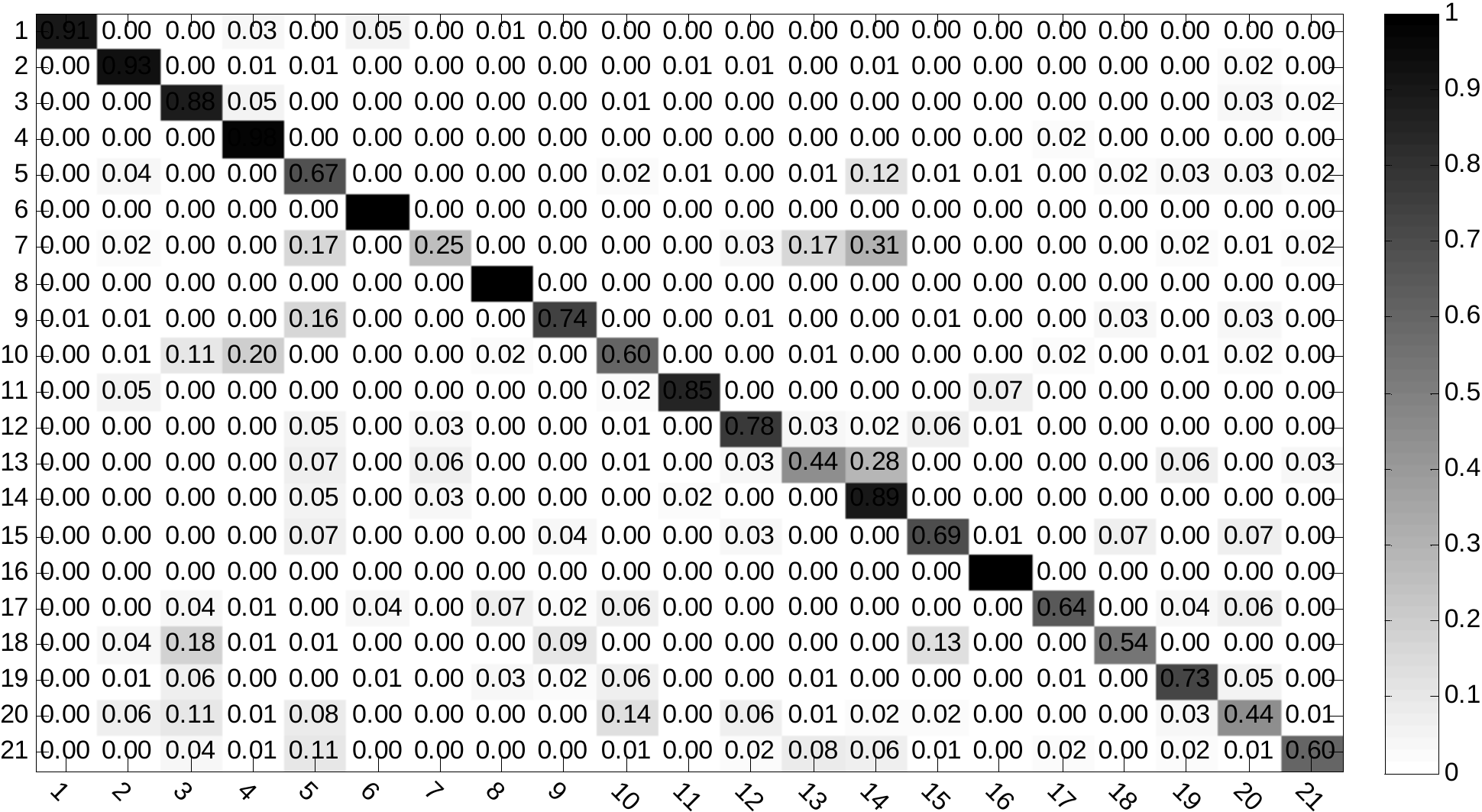}}
   \subfigure[]{\label{CM_21_50tr} \includegraphics[width=.45\linewidth]{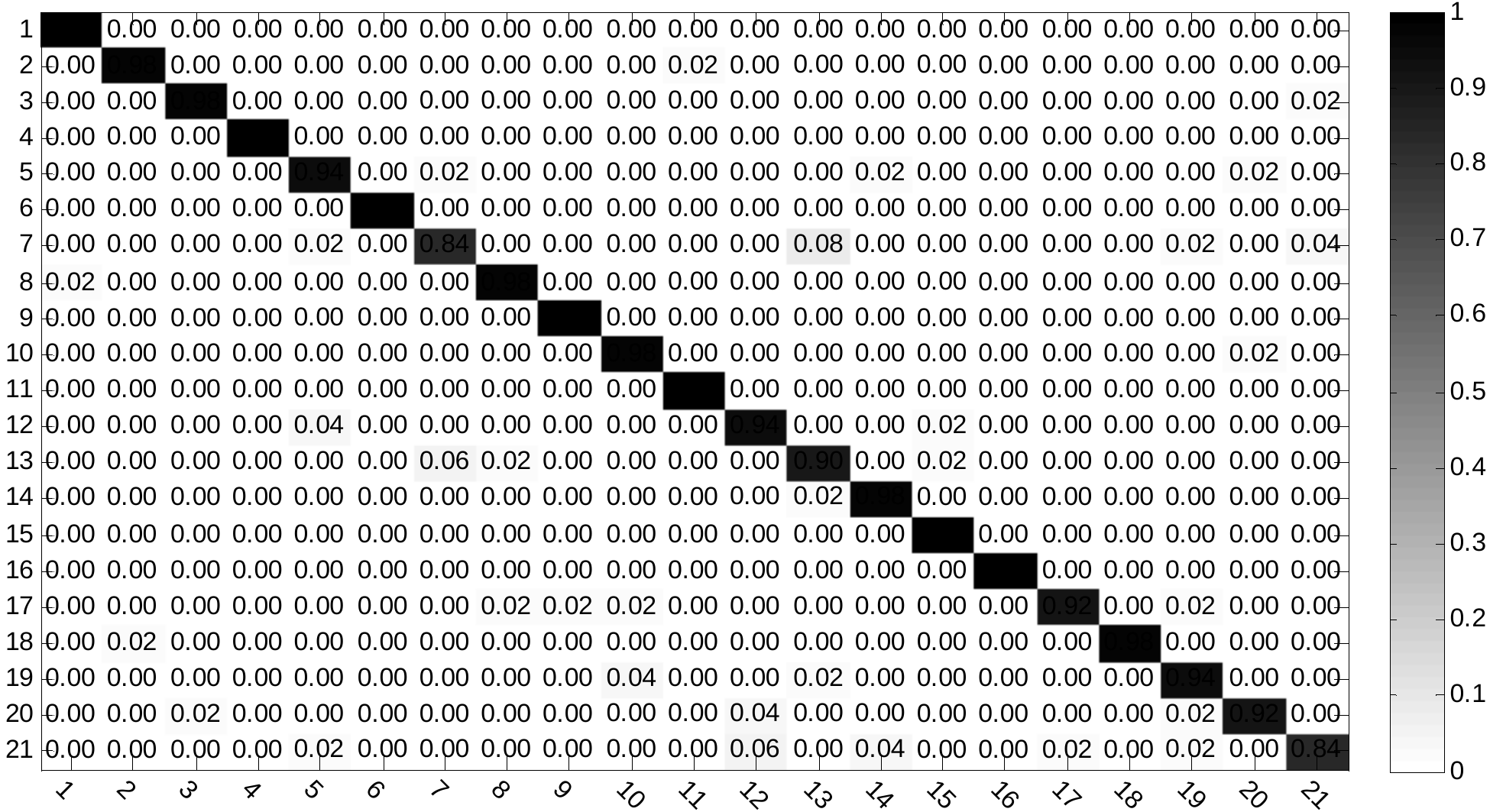}}
\end{center}
\caption{Confusion matrix of SPP-net+MKL with conv5+spp features under \subref{CM_21_5tr} 5 training samples and \subref{CM_21_50tr} 50 training samples in each class on 21-Class Satellite Scene dataset. The rows and columns of the matrix denote actual and predicted classes, respectively. The class labels are assigned as follows: 1 = Agricultural, 2 = Airplane, 3 = Baseball diamond, 4 = Beach, 5 = Buildings, 6 = Chaparral, 7 = Dense residential, 8 = Forest, 9 = Freeway, 10 = Golf course, 11 = Harbor, 12 = Intersection, 13 = Medium residential, 14 = Mobile home park, 15 = Overpass, 16 = Parking lot, 17 = River, 18 = Runway, 19 = Sparse residential, 20 = Storage tanks, and 21 = Tennis court.}
\label{confusion_matrix_21}
\end{figure}

\begin{figure}
  \centering
  \includegraphics[scale = 0.5]{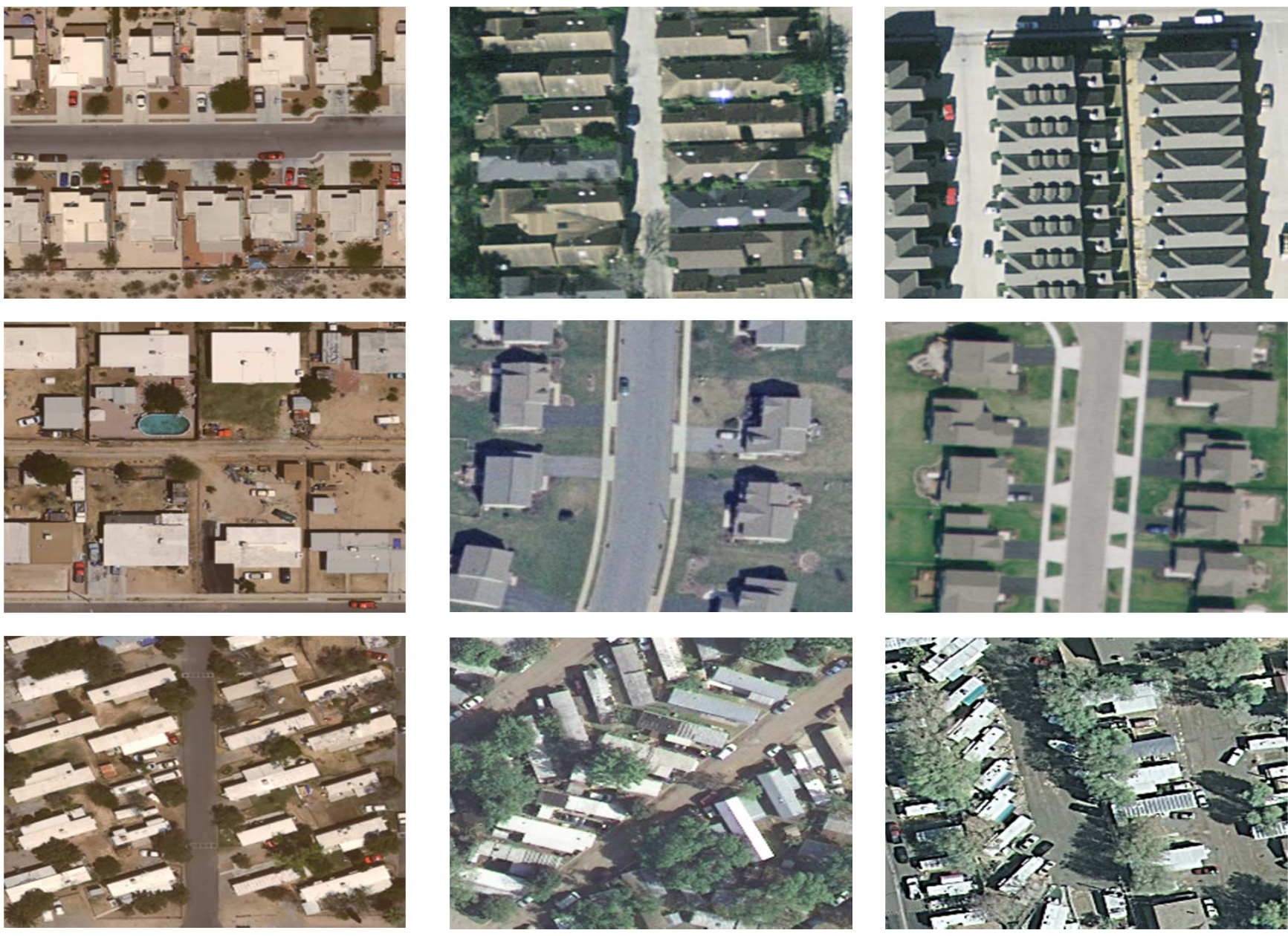}\\
  \caption{Some misclassified images in 21-Class-Land-Use dataset. Top: Misclassified images in \textit{dense residential} class. Middle: Some image examples of \textit{medium residential}. Bottom: Some image examples of \textit{mobile home park}.  }\label{classified_images_21}
\end{figure}

To demonstrate the superiority of the proposed method, we compare with several state-of-the-art approaches, including SSEP \cite{yang2015}, Partlets-based method \cite{cheng2015}, SC+Pooling \cite{cheriyadat2014}, BOVW \cite{yang2011}, SPCK++ \cite{yang2011}, SPMK \cite{lazebnik2006}, MKL \cite{cusano2014} and UFL \cite{hu2014}. The classification results with different number of training samples are shown in
table \ref{classification_comparision_21}, where `$-$' denotes there are no experiments. From this table, we can observe that SPP-net with the best single scale feature achieves higher accuracies than all comparison methods.
This implies that deep learning method learns more powerful features. Besides, the combination of multi-scale deep features further improves the performance. Specifically, SPP-net+MKL boosts the performance dramatically by 15\%, 8\% and 6\% in comparison with the existing best results when the number of training samples are 5, 50 and 80, respectively. To the best of our knowledge, these results are the best on this data set, which adequately show the superiority of our proposed method. In addition, we also compare SPP-net+MKL with two recently proposed state-of-the-art approaches by evaluating the accuracy in each class, which is shown in
Fig.$~$\ref{each_class_acc_21}. From Fig.$~$\ref{each_class_acc_21}(a), we observe that SSEP gets a little better performance than SPP-net+MKL in 6 classes. This is because SSEP method takes advantage of sampling technique to indirectly increase the number of training samples while SPP-net+MKL only uses the given training samples. Nevertheless, SPP-net achieves higher accuracies in the rest of 15 classes. Similarly, Fig.$~$\ref{each_class_acc_21} (b) demonstrates that SPP-net+MKL obtains higher performance in 19 classes compared to Partlets-based method in \cite{cheng2015}. For further analysis of the classification results achieved by SPP-net+MKL, we use confusion matrices shown in Fig.$~$\ref{confusion_matrix_21} to illustrate one of the results in ten experiments when the number of training samples is 5 and 50, respectively. The $i$th row and $j$th column element in confusion matrix denotes the rate of test samples from the $i$th class classified to the $j$th class. In the case of 5 training samples as shown in Fig.$~$\ref{confusion_matrix_21} (a), the most difficult classes to discriminate contain \textit{dense residential, Runway, medium residential} and \textit{storage tanks} whose accuracies are all lower than 60\%. For instance, as shown in Fig.~\ref{classified_images_21}, the \textit{dense residential} is easily misclassified as \textit{mobile home park} and \textit{medium residential} since they share similar building structures. However, when the number of training samples increases to 50, the accuracies of these classes improve significantly. This indicates that the number of training samples is a key factor for SPP-net+MKL.

\subsection{19-Class Satellite Scene dataset}
\subsubsection{\textbf{Data description and experimental setup}}
The second dataset is composed of 19 classes of scenes, including \textit{airport, beach, bridge, commercial area, desert, farmland, football field, forest, industrial area, meadow, mountain,
park, parking, pond, port, railway station, residential area, river} and \textit{viaduct}. Each class has 50 images, with size of $600\times600$ pixels. Such images are extracted from very large satellite images on Google Earth. Similar to 21-Class-Land-Use dataset, the original images are warped to three different scales: $128\times128$, $192\times192$ and $256\times256$.
We construct two experiments. The first one is randomly choosing 5 images from each class as the training set, and the rest images are used as the testing set, following \cite{sheng2012}\cite{yang2015}\cite{cusano2014}. The second experiment randomly chooses 25 images as the training set and the rest as the testing set, following \cite{sheng2012}.
All experiments are repeated 10 times with different training/testing split to obtain stable results. The final performance is reported as the mean and standard deviation of the results from 10 runs.

\subsubsection{\textbf{Each layer performance}}
\begin{figure}
\begin{center}
   \subfigure[]{\label{19_128} \includegraphics[width=.3\linewidth]{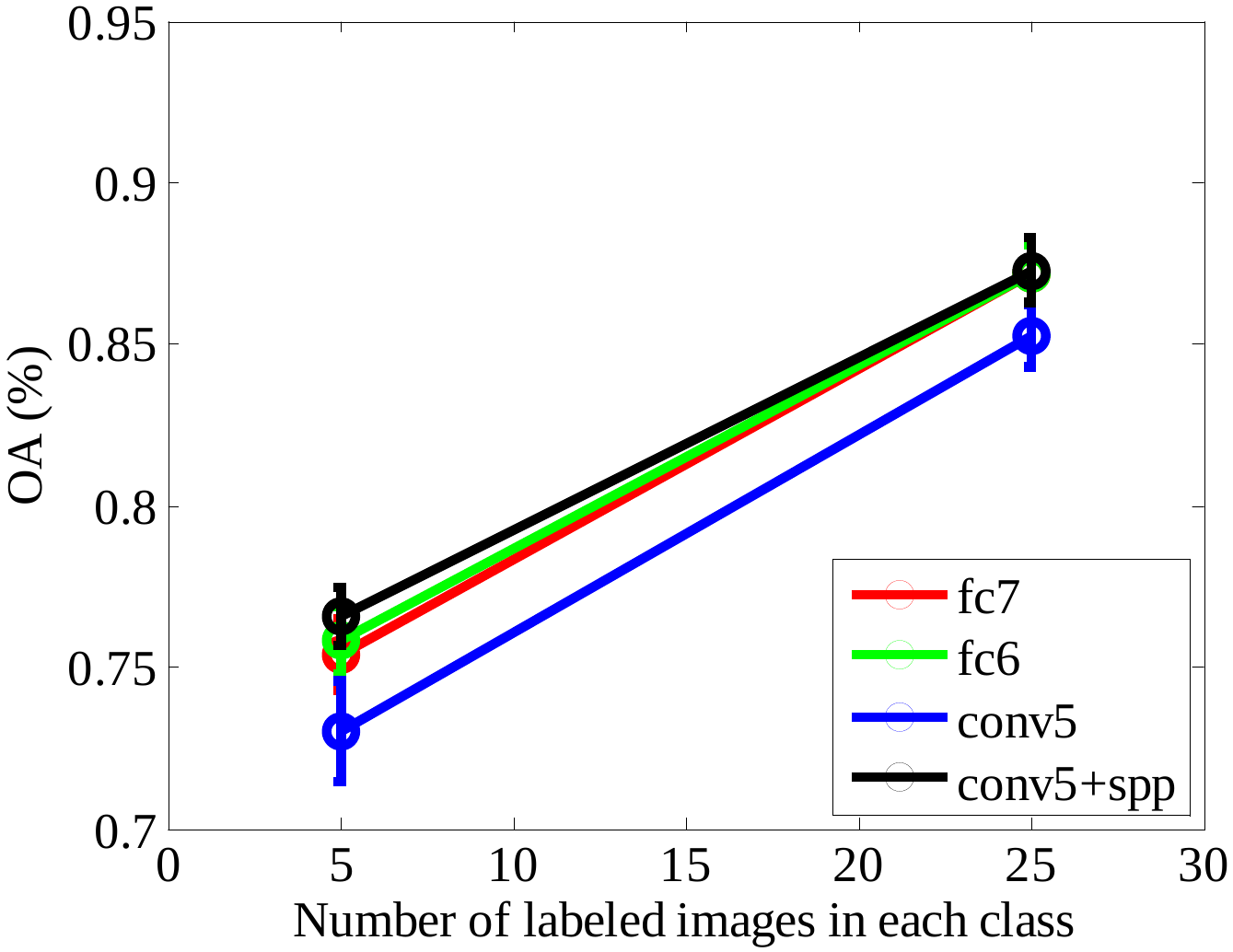}}
   \subfigure[]{\label{19_192} \includegraphics[width=.3\linewidth]{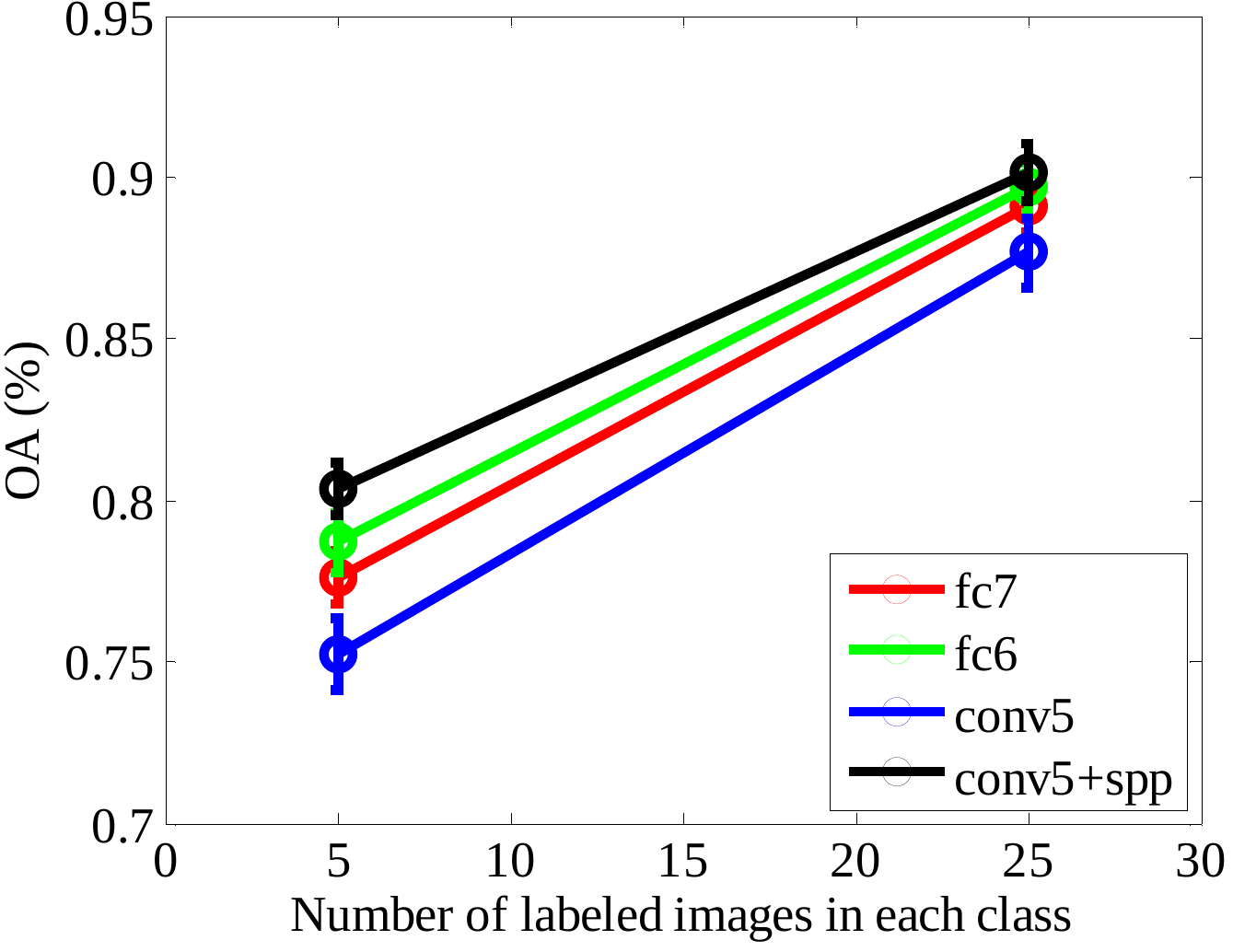}}
   \subfigure[]{\label{19_256} \includegraphics[width=.3\linewidth]{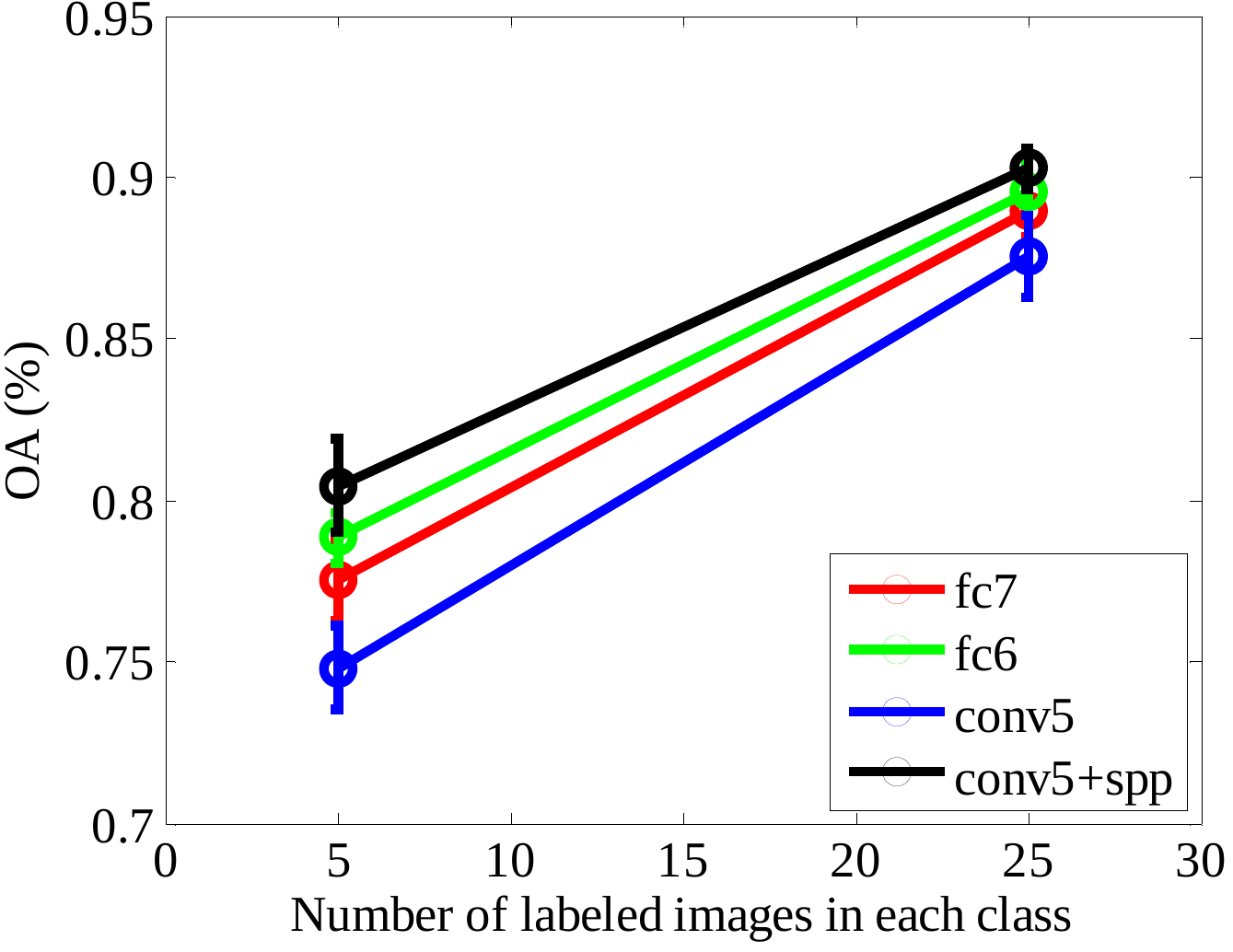}}
\end{center}
\caption{OAs and standard deviations of SPP-nets at three different scales under different number of training samples on 19-Class Satellite Scene dataset. \subref{19_128} $128\times128$ scale. \subref{19_192} $192\times192$ scale. \subref{19_256} $256\times256$ scale.}
\label{each_scale_results_19}
\end{figure}


\begin{figure}
  \centering
  \includegraphics[scale=0.5]{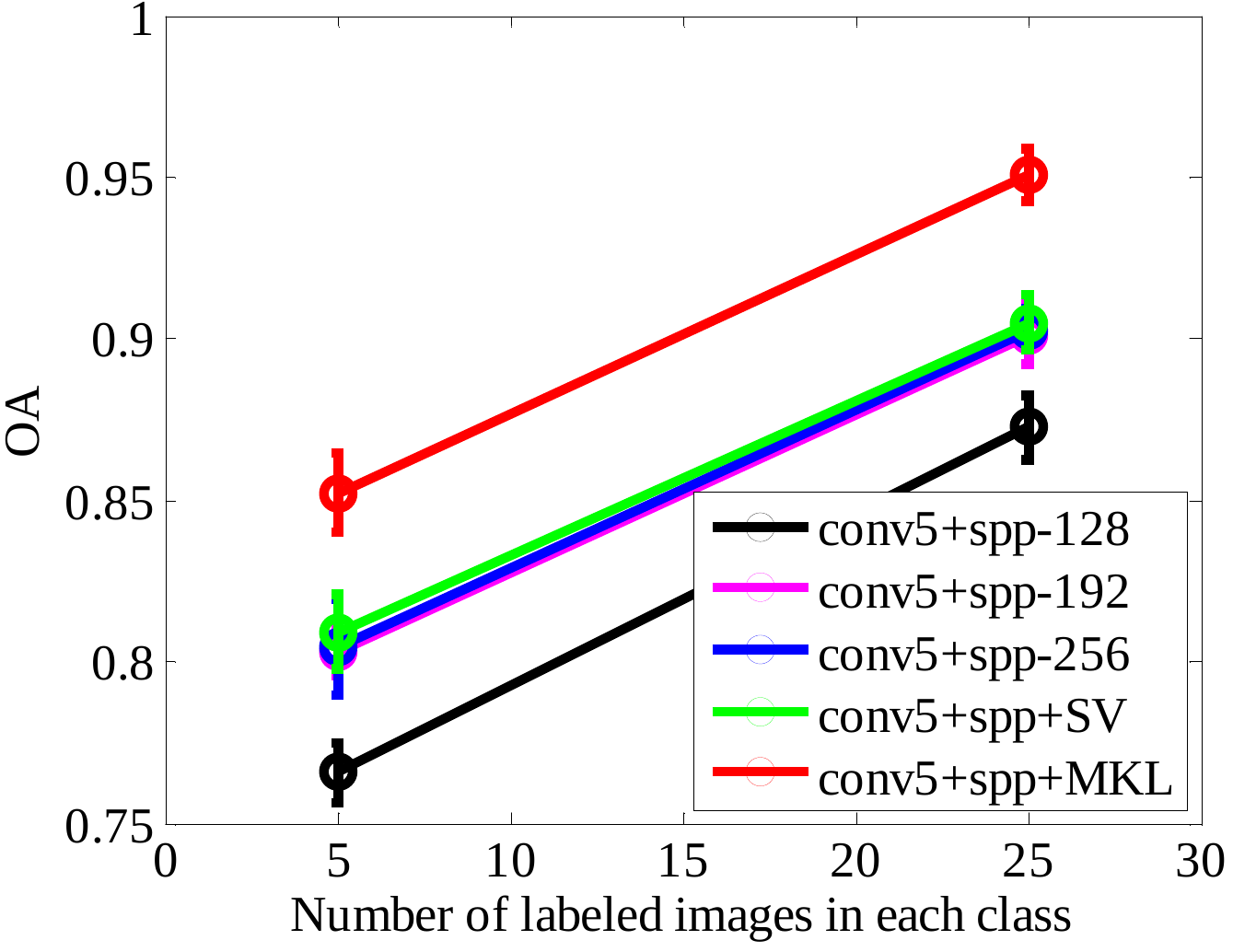}\\
  \caption{OAs and standard deviations of MKL versus SV and the single scales using conv5+spp features under different number of training samples on 19-Class Satellite Scene dataset.}\label{MKL_results_19}
\end{figure}

\begin{table}
  \centering
   \caption{ OAs (\%) and standard deviations of SPP-nets with different layer features under different number of training samples on 19-Class Satellite Scene dataset.}\label{classification_results_19}
  \begin{tabular}{|c|c|c|c|c|c|}
    \hline
    Scales & Numbers  & conv5 & conv5+spp  & fc6  & fc7\\
    \hline
    \multirow{2}{*}{$128\times128$} & 5  & $73.01\pm1.56$ & $\mathbf{76.56\pm0.92}$  & $75.82\pm1.03$ & $75.39\pm1.12$ \\
    \cline{2-6}
     & 25  & $85.22\pm0.97$ & $\mathbf{87.26\pm1.00}$ & $87.16\pm0.89$  & $87.16\pm0.89$ \\
    \cline{2-6}
    \hline
        \multirow{2}{*}{$192\times192$} & 5  & $75.25\pm1.10$ & $\mathbf{80.34\pm0.80}$  & $78.69\pm0.92$ & $77.60\pm0.85$\\
    \cline{2-6}
     & 25  & $87.68\pm1.08$ & $\mathbf{90.13\pm0.90}$  & $89.66\pm0.77$ & $89.12\pm0.85$\\
    \cline{2-6}
    \hline
    \multirow{2}{*}{$256\times256$} & 5  & $74.80\pm1.31$ & $\mathbf{80.46\pm1.47}$  & $78.84\pm0.80$ & $77.53\pm1.25$\\
    \cline{2-6}
     & 25 & $87.54\pm1.25$ & $\mathbf{90.27\pm0.64}$  & $89.57\pm0.76$ & $88.94\pm0.78$ \\
    \cline{2-6}
    \hline
  \end{tabular}
\end{table}

\begin{table}
  \centering
  \caption{The detailed classification results comparison between single scale features and two different
multi-scale feature fusion methods on 19-Class Satellite Scene dataset.}\label{MKL_results_19_table}
  \begin{tabular}{|c|c|c|}
     \hline
     Numbers & 5 & 25 \\
     \hline
     Conv5+spp-128 & $76.56\pm0.92$ & $87.26\pm1.00$ \\
     \hline
     Conv5+spp-192 & $80.34\pm0.80$ & $90.13\pm0.90$ \\
     \hline
     Conv5+spp-256 & $80.46\pm1.47$ & $90.27\pm0.64$ \\
     \hline
     Conv5+spp+SV & $80.92\pm1.16$ & $90.48\pm0.87$ \\
     \hline
     Conv5+spp+MKL & $\mathbf{85.22\pm1.22}$ & $\mathbf{95.07\pm0.79}$ \\
     \hline
   \end{tabular}

\end{table}

\begin{figure}[htp]
  \centering
  \includegraphics[scale = 0.5]{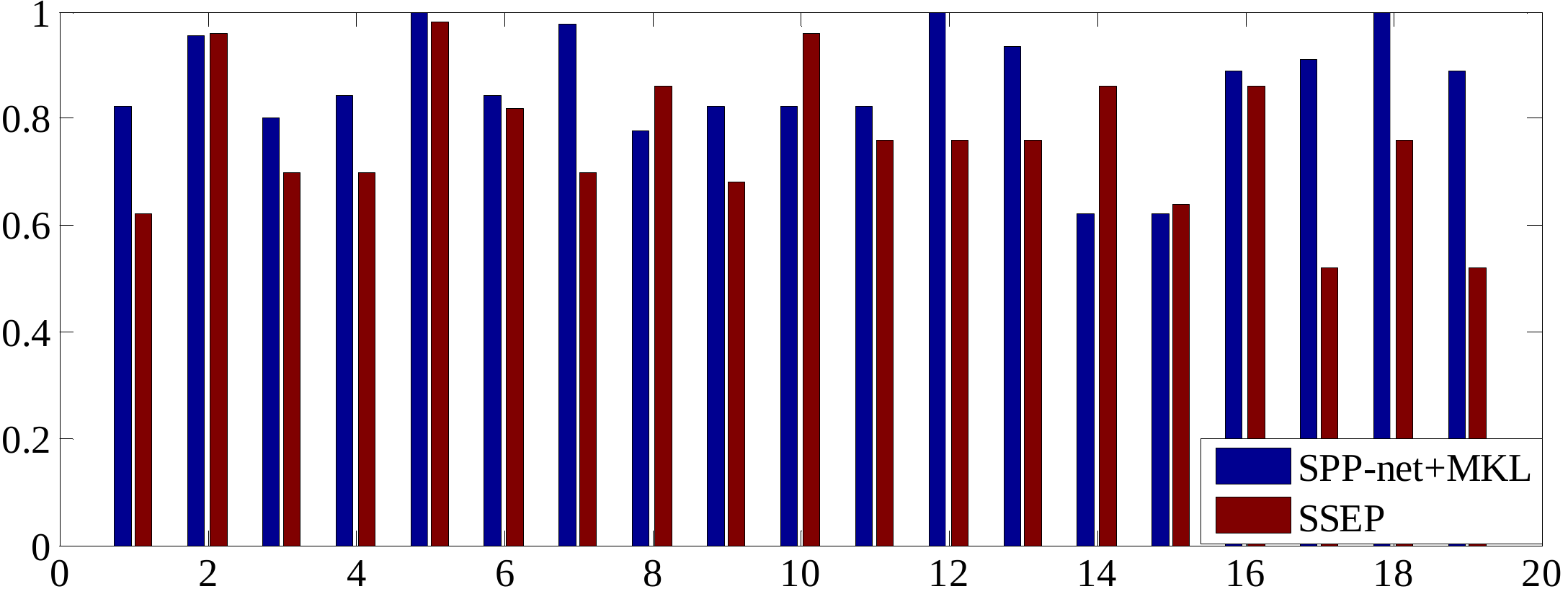}\\
  \caption{Each class accuracy comparison between SPP-net+MKL and SSEP in \cite{yang2015} using 5 training samples.}\label{each_class_acc_19}
\end{figure}

Similar to the first dataset, we evaluate the effect of different feature layers on the final performance. Fig.$~$\ref{each_scale_results_19} shows the classification results at three different scales using conv5, conv5+spp, fc6 and fc7 features. From this figure, we can observe that conv5+spp achieves the highest OAs as well as in the first dataset compared to the other three features, which is also demonstrated in
Table$~$\ref{classification_results_19}. Besides, we observe that the OAs on this dataset are higher than that on the first dataset under the same number of training samples. This is because this dataset is easier to discriminate and the number of testing set is smaller than that in the first dataset.
Fig.$~$\ref{MKL_results_19} and Table$~$\ref{MKL_results_19_table} compare the performance between single scale conv5+spp features and two multi-scale fusion methods. Obviously, the performance of SV is only a little better than that of $192\times192$ and $256\times256$ scales. However, MKL displays significant improvements in comparison with SV, which confirms the effectiveness of MKL fusion method.

\subsubsection{\textbf{Comparison with the state-of-the-arts}}
\begin{figure}[htp]
\begin{center}
   \subfigure[]{\label{CM_19_5tr} \includegraphics[width=.47\linewidth]{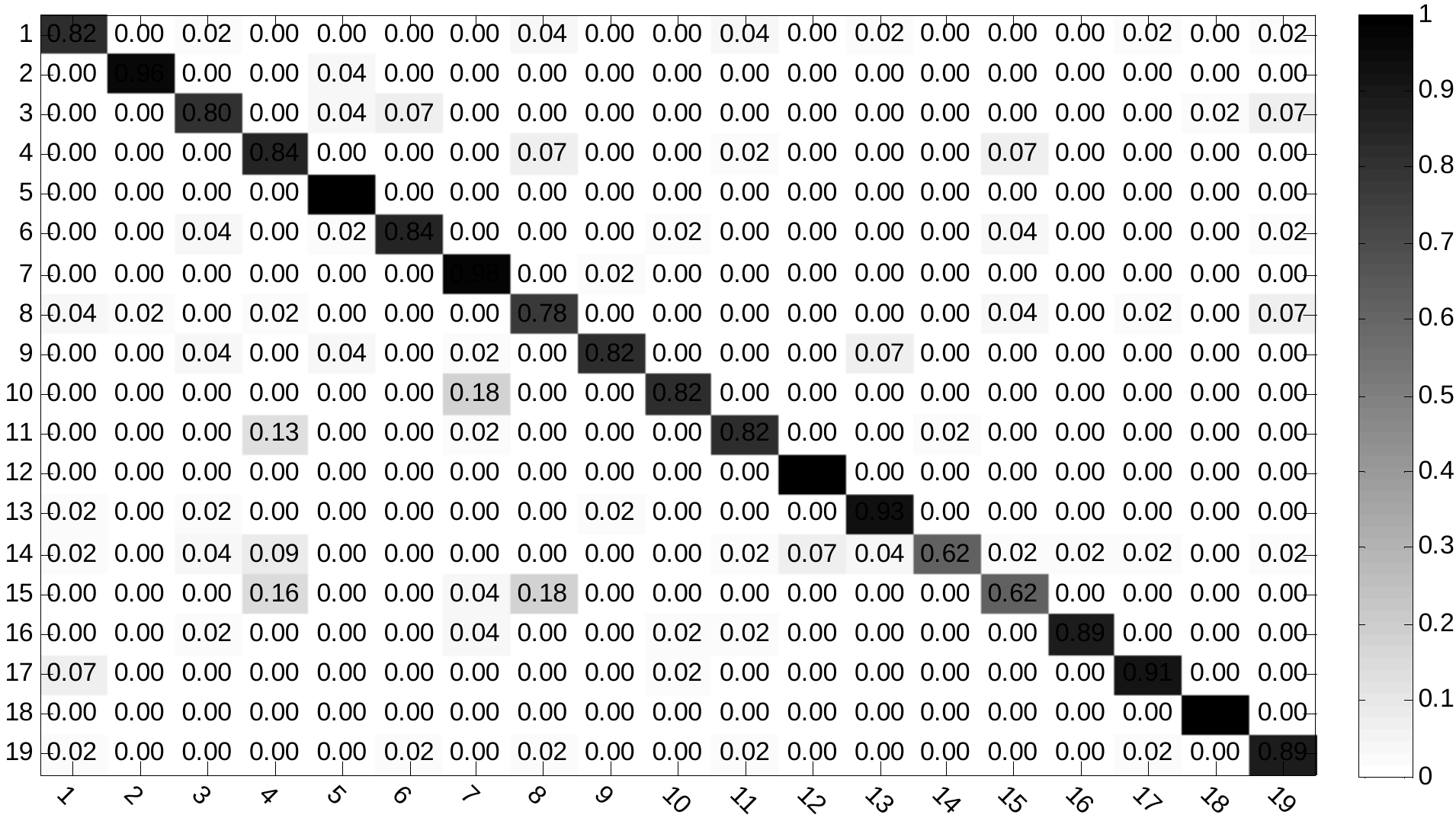}}
   \subfigure[]{\label{CM_19_25tr} \includegraphics[width=.47\linewidth]{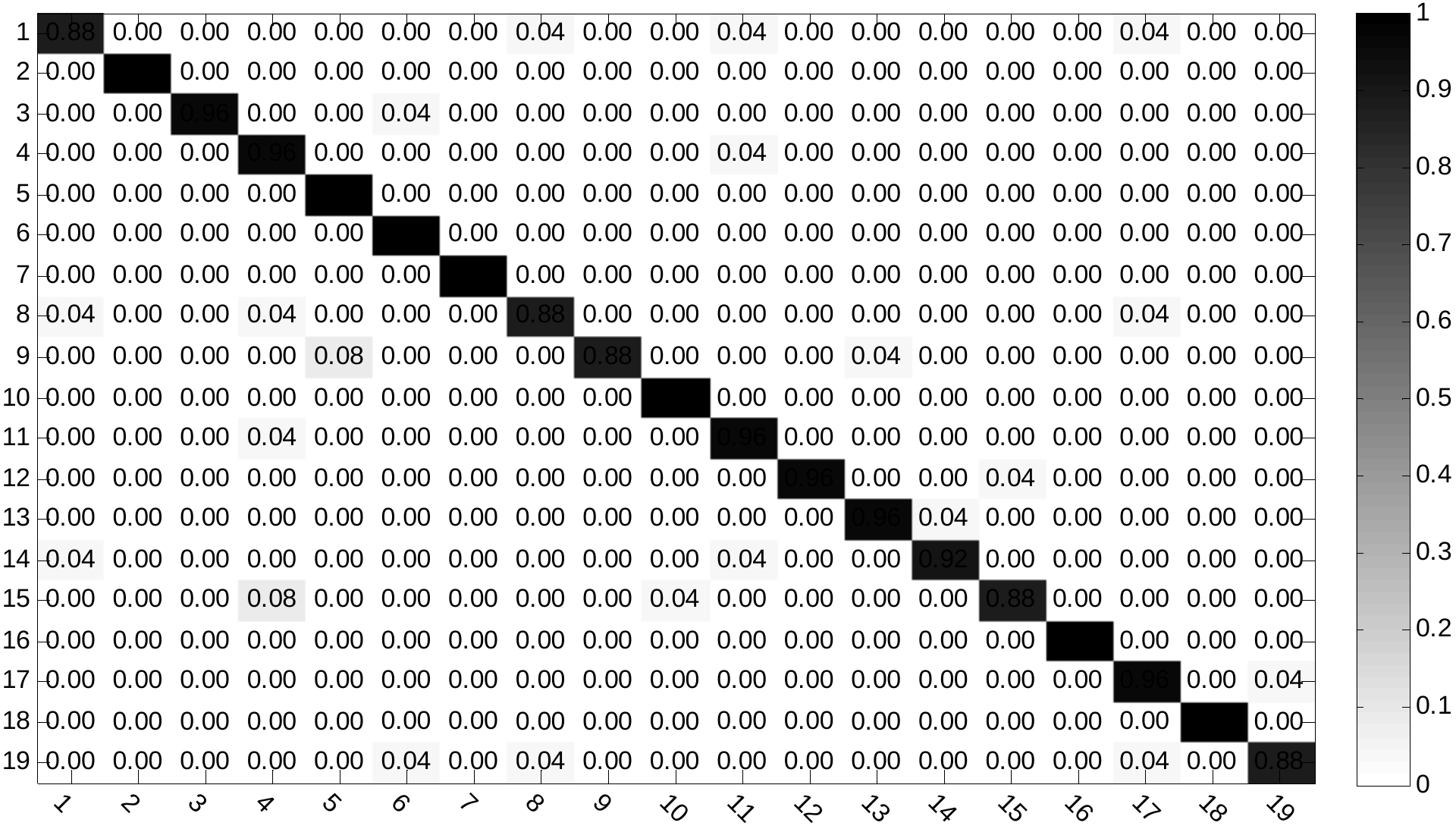}}
\end{center}
\caption{Confusion matrix of SPP-net+MKL with conv5+spp features under \subref{CM_19_5tr} 5 training samples and \subref{CM_19_25tr} 25 training samples in each class on 19-Class Satellite Scene dataset. The rows and columns of the matrix denote actual and predicted classes, respectively. The class labels are assigned as follows: 1 = Airport, 2 = Beach, 3 = Bridge, 4 = Commercial area, 5 = Desert, 6 = Farmland, 7 = Football field, 8 = Forest, 9 = Industrial area, 10 = Meadow, 11 = Mountain, 12 = Park, 13 = Parking, 14 =  Pond, 15 = Port, 16 = Railway station, 17 = Residential area, 18 = River, 19 = Viaduct.}
\label{confusion_matrix_19}
\end{figure}

\begin{table}[htp]
  \centering
    \caption{Overall classification accuracy (\%) comparison on the 19-Class Satellite Scene dataset.}\label{classification_comparision_19}
  \begin{tabular}{|c|c|c|}
     \hline
     Numbers & 5 & 25 \\
     \hline
     SSEP \cite{yang2015}  & $73.82\pm1.52$ & $-$ \\
     \hline
     SCMF \cite{sheng2012} & 78.32 & 90.05 \\
     \hline
     MKL \cite{cusano2014} & $67.32\pm2.90$ & $-$ \\
     \hline
     SPP-net               & $80.46\pm1.47$ & $90.27\pm0.64$  \\
     \hline
     SPP-net+SV               & $80.92\pm1.16$ & $90.48\pm0.87$  \\
     \hline
     SPP-net+MKL               & $\mathbf{85.22\pm1.22}$ & $\mathbf{95.07\pm0.79}$  \\
     \hline
   \end{tabular}
\end{table}

In order to comprehensively analyze the superiority of the proposed method, we compare it with three state-of-the-art approaches ever tested on this dataset. They are SSEP \cite{yang2015}, SCMF \cite{sheng2012} and MKL \cite{cusano2014}. The comparison results are illustrated in
Table$~$\ref{classification_comparision_19}, from which we can observe that the proposed SPP-net+MKL significantly improves the accuracy from 78.32 to 85.22 and 90.05 to 95.07 when the numbers of training samples are 5 and 25, respectively. Besides, we compare each class accuracy with the latest approach SSEP in \cite{yang2015}. SPP-net achieves higher accuracies in 14 classes as shown in Fig.$~$\ref{each_class_acc_19}. To further analyze the classification results of SPP-net+MKL, we demonstrate the confusion matrices of one of the results in ten experiments when the number of training samples is 5 and 25 in Fig.$~$\ref{confusion_matrix_19}. It is observed that \textit{ port} is easily to be misclassified as \textit{forest} due to the small number of training samples (i.e., 5).  Similar results appear in \textit{meadow}. Fortunately, as the number of training samples increases to 25, both of the accuracies improve dramatically.

\section{Conclusion}
This paper proposed to automatically extract multi-scale deep features from the satellite images by using SPP-net. This net comprises five convolutional layers and two fully-connected layers where the last convolutional layer is followed by the spatial pyramid pooling operator. It is well known that the performance of deep models heavily depends on the large number of training samples, while only hundreds of samples are available in most of the satellite image classification cases. Therefore, we focused on solving the problem of training multiple effective SPP-nets simultaneously. To this end, we pre-trained the DCNN model by using auxiliary ImageNet dataset, which is different from satellite images, and then transferred the parameters in the five convolutional layers to the SPP-nets. Finally, the fully-connected layers of each SPP-net were fine-tuned by their corresponding training samples. It is of great interest to see that this training approach leads to very promising classification results that outperform the existing best results on the same data sets. Furthermore, a multiple kernel learning method was adopted to fuse the multi-scale deep features. The experiments on two classical satellite datasets have demonstrated that the proposed method dramatically improves the classification results compared with several state-of-the-arts.

\bibliography{IEEEfull,spp}
\bibliographystyle{IEEEbib}
\end{spacing}
\end{document}